\documentclass[final,5p,times,twocolumn,authoryear]{elsarticle}

\usepackage{framed,multirow}

\usepackage{amssymb}
\usepackage{latexsym}

\usepackage{url}
\usepackage{xcolor}

\usepackage{hyperref}

\definecolor{newcolor}{rgb}{.8,.349,.1}

\usepackage[utf8]{inputenc}
\usepackage{amsmath, amssymb}
\usepackage{graphicx}
\usepackage[]{setspace}
\usepackage{xcolor}
\usepackage{multirow}
\usepackage{caption}
\usepackage{subcaption}
\usepackage{svg}

\journal{a journal}

\begin{document}

\begin{frontmatter}

\title{Prompt Tuning for Parameter-efficient Medical Image Segmentation\tnoteref{tnote1}}%

\author[1]{Marc Fischer\corref{cor1}}
\cortext[cor1]{Corresponding author}
\ead{marc.fischer@iss.uni-stuttgart.de}
\author[1]{Alexander Bartler} 
\author[1]{Bin Yang}
\fntext[footnote_abbreviations]{Abbreviations: prompt-able UNet (PUNet), prompt-able shifted window (PSWin), prompt-able multi-head attention (PMA), contrastive prototype assignment (CPA)}

\address[1]{Institute of Signal Processing and System Theory, University of Stuttgart, 70550 Stuttgart, Germany}

\begin{abstract}
Neural networks pre-trained on a self-supervision scheme have become the standard when operating in data rich environments with scarce annotations. As such, fine-tuning a model to a downstream task in a parameter-efficient but effective way, e.g. for a new set of classes in the case of semantic segmentation, is of increasing importance. In this work, we propose and investigate several contributions to achieve a parameter-efficient but effective adaptation for semantic segmentation on two medical imaging datasets. Relying on the recently popularized prompt tuning approach, we provide a prompt-able UNet (PUNet) architecture, that is frozen after pre-training, but adaptable throughout the network by class-dependent learnable prompt tokens. We pre-train this architecture with a dedicated dense self-supervision scheme based on assignments to online generated prototypes (contrastive prototype assignment, CPA) of a student teacher combination alongside a concurrent segmentation loss on a subset of classes. We demonstrate that the resulting neural network model is able to attenuate the gap between fully fine-tuned and parameter-efficiently adapted models on CT imaging datasets. As such, the difference between fully fine-tuned and prompt-tuned variants amounts to only $3.83$ pp for the TCIA/BTCV dataset and $2.67$ pp for the CT-ORG dataset in the mean Dice Similarity Coefficient (DSC, in \%) while only prompt tokens, corresponding to $0.85\%$ of the pre-trained backbone model with 6.8M frozen parameters, are adjusted. The code for this work is available on \url{https://github.com/marcdcfischer/PUNet}.
\end{abstract}


\end{frontmatter}


\section{Introduction}
With an ever increasing amount of radiological images to analyze, computer aided diagnosis (CAD) has become an integral part of medical studies. A basic processing block of many analyses is the semantic segmentation of medical imaging data. A plethora of approaches and neural network architectures has been proposed. Hereby, deep learning (DL) not only provides fast, but also robust and reproducible results, given sufficient quality of and annotations for the imaging data.

Commonly known, medical images are hard to segment due to its prevalence for inhomogeneities and variabilities. As such, many architectural advancements have been introduced for convolutional neural networks (CNN, \cite{khan2020survey}) that progressively increased the robustness and performance of these networks. Recently, architectures that rely predominantly on transformer blocks \citep{vaswani2017attention} instead of traditional convolutions have shown a more broad applicability and superior performance. Medical variants have been proposed \citep{tang2022self, cao2021swin}, which follow the popular UNet encoder-decoder structure \citep{ronneberger2015u} and allow for excellent semantic segmentation.

In addition, the generalization capabilities of semantic segmentation approaches to unseen medical imaging data was long constrained by the scarcity and quality of annotations. Numerous schemes have been explored to alleviate this circumstance \citep{tajbakhsh2020embracing}. Recently, self-supervision by contrastive learning \citep{oord2018representation, chen2020simple} has become incremental to leverage the abundance of (imaging) data, while keeping the requirement for the amount of annotated data low. Neural network weights are identified by a dedicated self-supervised pre-training scheme and serve as a basis for further downstream adaptation. A simple way to perform such an adaptation to the task at hand is to fully fine-tune the model by adjusting all its parameters. However, such a procedure incurs high memory and training time costs. 

Despite these foundational contributions, self-supervised semantic segmentation models are often fine-tuned by adding an entirely new and task dependent decoder for every new encountered task (e.g. in the case of a new set of unseen classes). As such, the resulting end-to-end models are trained once for a specific combination of training data and available annotations. Inspired by recent successes in the natural language processing (NLP) community \citep{liu2021pre, li2021prefix, lester2021power}, we instead consider adapting neural network architectures by inserting additional tokens, called prompts, besides the input data. Tuning these prompts allows for the conditioning of pre-trained and subsequently frozen (language) models to perform specific downstream tasks. Hereby, learned continuous prompt tokens (vectors) are prepended to the (embedded) input. The prompt tokens are updated by gradient decent in the backward pass like regular trainable neural network weights. However, they remain part of the input space and can be replaced or relearned for a new task while leaving the (frozen) backbone network task agnostic. This allows for a high degree of parameter sharing between tasks and thus low resulting memory costs when a model is extended to a new tasks.

From a practical standpoint, training only one architecture and adapting it in a parameter efficient way, opens up new possibilities on how to use the models in the medical context. We envision, only one (general purpose) architecture to be trained, which can be stored in a suitable (remote) location. On-premise, it can be adapted to the task of interest. As such, only minimal amounts of parameters have be stored, which would even allow for e.g. subject specific model tuning, which could aid in longitudinal studies. In addition, the private patient data remains on the location where the model was adapted on. In addition, task capabilities of a model could be extended and shared by publishing new prompt tokens. 

To enable such a prompt tuning scheme for semantic segmentation on medical imaging data, we revisit aforementioned advancements and adapt them with the goal to facilitate task adaptation of a pre-trained but frozen model with minimal parameter adjustments. To this end, we
\begin{itemize}
    \item introduce a deeply prompt-able encoder-decoder architecture (prompt-able UNet, PUNet) that can incorporate additional class-dependent prompt tokens to achieve dense binary and multi-class segmentation, 
    \item contribute architectural components comprising prompt-able shifted window (PSWin) blocks, a heterogeneous bias score generation within the attention scheme, and a weighted similarity aggregation to enable token-dependent class predictions,
    \item propose a contrastive pre-training scheme specifically designed for dense self-supervision by soft assignments to online generated prototypes to establish anatomical representations while circumventing a hard separation of the contrastive attraction and repulsion,
    \item show that "prompting" of the pre-trained and frozen architecture by non-frozen (learned) prompt tokens is sufficient for adaptation to a segmentation downstream task on medical imaging data,
    \item leverage our assignement-based self-supervision scheme to enable the concurrent application of a prompt-dependent segmentation supervision in the pre-training phase, further reducing the performance gap between fully fine-tuned and efficiently adapted variants.
\end{itemize}

\section{Related work}

The work builds upon multiple recent developments. Notable mentions include the reliance on large models, the possibility for prompt tuning as downstream task adaptation, as well as recent advancements in using transformer blocks, but also self-supervised learning, and alternative pre-training strategies known from meta learning. Most notable works in these fields with respect to natural and medical images will be briefly covered.

\subsection{Neural Network Architecture}
Architectural design underwent a change from convolutional neural networks, especially suited for image data, to general-purpose transformer-based architectures \citep{vaswani2017attention} relying on attention layers. Originating in the NLP community, this design was subsequently applied to natural images by means of the Vision Transformer (ViT) \citep{dosovitskiy2020image}. Like its language model pendant, the vision variant can be scaled to billions of parameters \citep{zhai2022scaling}. In the following, transformer-based architectures have gained significant interest for image classification \citep{liu2021swin}, but also semantic segmentation \citep{strudel2021segmenter, zheng2021rethinking}, panoptic segmentation \citep{cheng2021per}, and have been applied successfully to medical images \citep{chen2021transunet, xie2021cotr, hatamizadeh2022unetr, tang2022self}. \cite{strudel2021segmenter} showed that a transformer named Segmenter, consisting of an encoder together with a joint transformer decoder for encoded content and output class tokens, can achieve superior results to popular convolutional architectures, such as DeepLabv3+ \citep{chen2018encoder}. Later, shifted window (SWin) blocks \citep{liu2021swin} were introduced that greatly reduced the memory costs by limiting the self-attention to local non-overlapping windows that are subsequently shifted. Thus, a linear complexity is maintained in comparison to the quadratic complexity induced by self-attention layers. As such, the application of the attention scheme becomes viable throughout the encoder and decoder on medical data \citep{cao2021swin} while keeping memory requirements low. Still, for medical data, the encoder-decoder form with skip connections, popularized by the UNet \citep{ronneberger2015u}, remained prevalent. As such, \cite{tang2022self} proposed a SWin UNet Transformer (SWinUNETR) with SWin transformer blocks in the encoder and convolutional layers in the decoder. \cite{cao2021swin} used a Shifted Window (SWin) UNet (Swin-UNet) with SWin transformer blocks in the encoder as well as the decoder.

\subsection{Fine-Tuning}
In recent years, model sizes have quickly grown from single digit millions \citep{he2016deep}, to hundreds of millions \citep{devlin2018bert}, and more recently billions of parameters \citep{chowdhery2022palm}. For very large architectures which are pre-trained on curated data at scale, one also speaks of foundation models \citep{bommasani2021opportunities}. These models allow for unprecedented generalization and few-shot adaptation capabilities. However, with the increase of model parameters, effective and parameter-efficient adaptation methods of the transfer learning field become mandatory for model tuning to respective downstream tasks. As such, over-fitting on small scale downstream datasets and expansive storage costs can be alleviated. Such fine-tuning approaches adjust or inject a small trainable subset into a larger pre-trained model and optimize those parameters on specific downstream tasks. Straightforward approaches include the replacement or addition of several output layers \cite{mahajan2018exploring}, ranging from a classification head for classification to entire decoders for dense predictions. It is also possible to continue optimizing the bias terms of all neural network layers, while keeping the remaining bulk of parameters fixed \citep{zaken2021bitfit}. Another parameter-efficient way is to include additional residual layers \citep{houlsby2019parameter}, called adapters, within existing blocks of layers. More sophisticated schemes achieve similar results by means of pruning based on sparse difference vectors \citep{guo2020parameter} or small additive side networks \citep{zhang2020side}. \cite{he2021towards} proposed a unified view that generalizes some of the aforementioned concepts. For natural images \citep{jia2022visual} evaluated a subset of these approaches for image classification and segmentation.  

Recently, a new paradigm called prompt tuning \cite{liu2021pre}, achieved great successes in the NLP community \citep{li2021prefix, lester2021power}. Motivated by this success, first approaches explored this adaptation scheme for natural images by directly adapting the pixel space \citep{bahng2022visual} or via Visual Prompt Tuning (VPT) by \cite{jia2022visual}, where prompts can be injected deeply into the blocks of a dedicated frozen vision backbone \citep{dosovitskiy2020image}. Naturally, sophisticated generation schemes for prompts can be envisioned. For example, \cite{he2022hyperprompt} explore a HyperPrompt framework by generating layer specific prompt tokens based on global prompts and small networks.

\subsection{Few-Shot Segmentation}
Few-shot segmentation has been investigated to provide a robust generalization performance in the presence of few annotated samples. Hereby, prototypical networks \citep{snell2017prototypical, wang2019panet}, an instance of meta learning, have been explored. These networks learn to predict an embedding in which prototypes of classes can be established as robust and discriminative representatives. Another avenue relies on supervised pre-training on an available dataset followed by a subsequent transfer learning. Yet, with respect to medical images, \cite{raghu2019transfusion} highlight issues for transfer learning of supervised models on natural data to medical imaging. However, for transfer learning between different medical datasets studies like \cite{chen2019med3d} have established the effectiveness of supervised pre-training schemes. Nontheless, their widespread adoption remains limited due to the inherent cost associated with providing expert annotations for 3D medical volumes at scale. We note that the main training process in VPT \citep{jia2022visual} also follows a supervised pre-training performed on another dataset. 

\begin{figure*}[h]
    \centering
    \includegraphics[width=1.\textwidth]{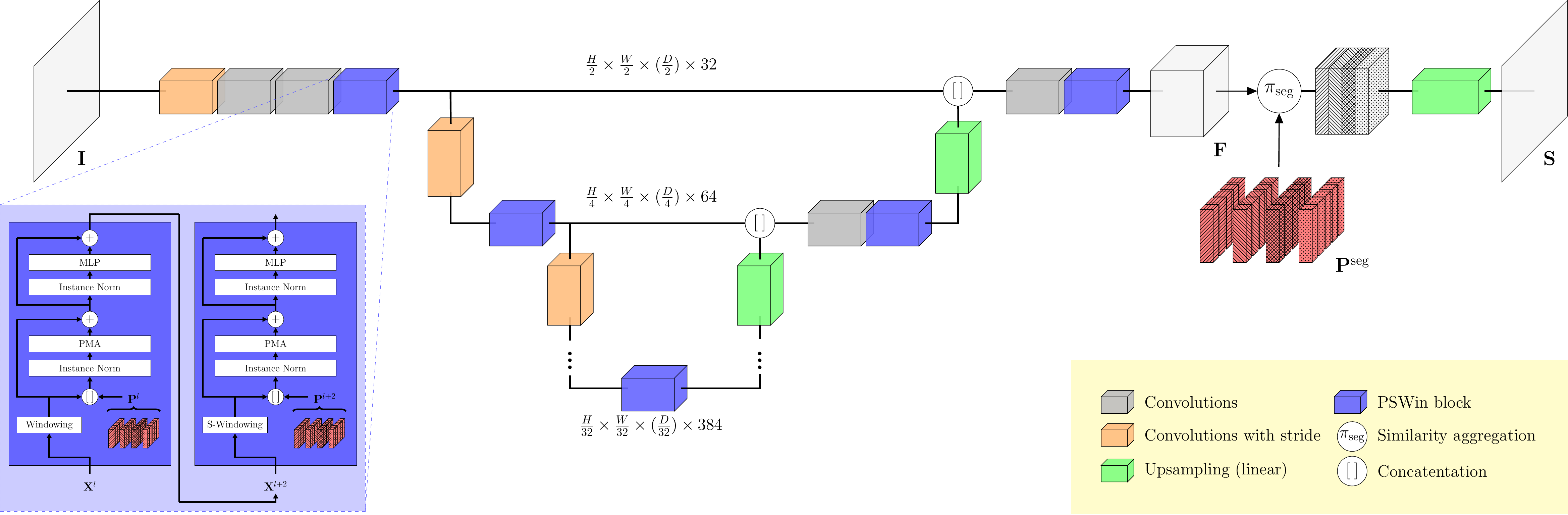}
    \caption{Schematic illustration of the proposed prompt-able UNet (PUNet). The network consists of an encoder with down-convolutions and a decoder with linear upsampling layers. A depth of 5 levels is chosen with 32, 64, 128, 256, 384 hidden channels $C$ for each respective level. Throughout the network prompt-able shifted window (PSWin) blocks are placed. These blocks incorporate prompt-able multi-head attention (PMA) layers. This enable the injection of prompt tokens $\mathbf{P}$ that can be learned in a downstream adaptation task. The decoder embedding $\mathbf{F}$ is further processed by a similarity aggregation $\pi_{\mathrm{seg}}$ together with prompt tokens $\mathbf{P}^{\mathrm{seg}}$ for the prediction of class probabilities in the segmentation map $\mathbf{S}$.}
    \label{fig:instructed_architecture}
\end{figure*}

\subsection{Self-supervised Representation Learning}
\label{sec_refs_self_sup}
In the medical field, self-supervision \citep{taleb20203d, azizi2021big, ghesu2022self} has become an integral part for the pre-training of models on the abundance of available (raw) data. Early approaches designed pre-text tasks, such as recombining jigsaw puzzles \citep{noroozi2016unsupervised}, solving a rubik's cube \citep{zhuang2019self}, predicting rotations \citep{gidaris2018unsupervised} or learning to reconstruct masked input content based on context by inpainting on natural \citep{pathak2016context} and medical \citep{chen2019self, haghighi2021transferable, zhou2021models} images. The consistent anatomy allows for learning of anatomical regions, which can be leveraged for, e.g. landmark localization \citep{yan2022sam} or segmentation \citep{chaitanya2020contrastive}. Hereby, the aforementioned vision transformers have been identified as excellent candidates for pre-training on natural images \citep{bao2021beit, chen2021empirical, caron2021emerging} as well as on medical images \citep{liu2021swin}. Combinations of the schemes are also possible. For example, \cite{tang2022self} used a combination of inpainting, contrastive learning and rotation prediction to pre-train a SWin transformer encoder.

Several sophisticated approaches proposed to use contrastive learning in combination with prototypes to achieve robust embeddings. Again, these prototypes are codes which are used as surrogate targets. Following the seminal work of \cite{oord2018representation} and their contrastive InfoNCE formulation, the prediction of direct similarities as popularized in \citep{chen2020simple} can be replaced by assignments to a surrogate \citep{li2020prototypical}. In many cases, a momentum-updated teacher student combination allows for a variety of applications and augmentations \citep{baevski2022data2vec}. Robust prototype codes can be either learned \citep{caron2020unsupervised}, drawn from support samples \citep{assran2021semi}, aggregated into a (compressed or quantized) queue \cite{dwibedi2021little}, generated by clustering \cite{yue2021prototypical}) or be derived by predicted embeddings from a target such as a quantized vector out of a non-augmented source image \cite{gidaris2021obow}. Recently, \cite{henaff2022object} relied on assignments to online generated dense masks by a k-means algorithm. These masks are in turn aggregated to object level prototypes and mapped to students as targets. An alternative approach has been considered by \cite{ouyang2020self}, who proposed the use of offline generated clusters of superpixels for local and global surrogate targets. In general, assignments to prototypes provide a method for generalization unmatched by straightforward reconstruction schemes.

\section{Methods}

\begin{figure*}[t]
    \centering
    \includegraphics[width=1.\textwidth]{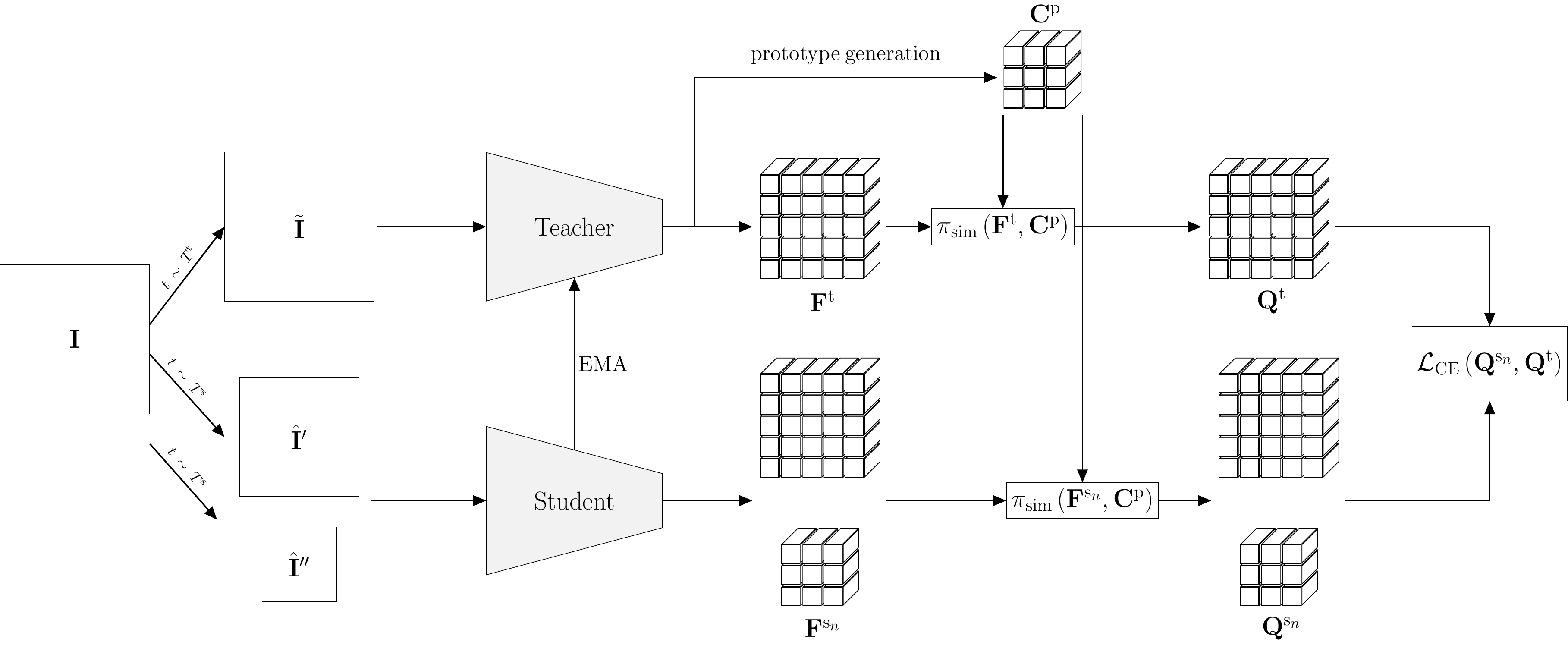}
    \caption{Input image slices $\mathbf{I}$ are augmented for a student teacher combination. Differently sized views $\hat{\mathbf{I}}'$, $\hat{\mathbf{I}}''$, with varying augmentations including partial masking, are passed to a student as well as a less augmented variant $\tilde{\mathbf{I}}$ to a teacher. The teacher network weights are kept up to date by an exponential moving average (EMA). Teacher output embeddings $\mathbf{F}^{\mathrm{t}}$ are further processed by an iterative clustering with spatially weighted assignments, which results in online generated prototypes $\mathbf{C}^{\mathrm{p}}$. Similarity assignments are calculated by means of a softmaxed cosine similarity $\pi_{\mathrm{sim}}$ with respect to predicted embeddings $\mathbf{F}^{\mathrm{t}}$ and $\mathbf{F}^{\mathrm{s}_n}$ for the $n$th student, as well as the prototype codes $\mathbf{C}^{\mathrm{p}}$. Derived assignments $\mathbf{Q}^{\mathrm{t}}$ and $\mathbf{Q}^{\mathrm{s}_n}$ are enforced to have similar contrastive prototype assignments (CPA) by a cross-entropy (CE) loss.}
    \label{fig:training_scheme}
\end{figure*}

The methodical contributions can be separated into three aspects: incorporated architectural components, a dense self-supervision scheme and the dedicated training scheme. In order to achieve an architecture capable of the aforementioned goals, we first introduce an UNet-like architecture in Section \ref{sec_punet} that can be prompted throughout the network. Hereby, prompt-able SWin blocks are presented in \ref{sec_pswin} which make use of recent advances of self-attention (transformer) layers. These are integrated in an encoder-decoder architecture suitable for medical images. Further adjustments include relative positional encodings as attention bias scores for the image content as well as learned encodings for the calculation of attention bias scores between prompts and image content. Together, they constitute the heterogeneous attention bias score introduced in Section \ref{sec_bias}. To keep the architecture flexible from start to end, a dedicated output layer is employed by a weighted similarity aggregation across tokens, in order to predict desired class probabilities (Section \ref{sec_aggregation}).

We propose a dense self-supervision scheme applied during pre-training by a contrastive assignment-based loss in Section \ref{sec_self}. In comparison to most common pre-training schemes, the whole backbone model is optimized due to the densely penalized output. It relies on a student teacher combination with two students, an exponential moving average (EMA)-updated teacher, and online generated prototype targets. In spite of altered variations of similar field of views (FOV), students are enforced to output similar assignments and thereby predicted representations as the teacher model (Section \ref{sec_cpa}. The online clustering based on soft assignments weighted by relative spatial distances to prototype seeds is established in \ref{sec_prototypes}. The initial prototype seeds are drawn from the teacher predictions itself. The scheme and its formulation are adequate for extension to a concurrent class-conditional supervision by prompt insertion. We propose to apply the prompts per batch element to enhance the adaptation of the learned embedding and thereby its manipulation by instructions during the subsequent downstream adaptation. Resulting variations of the training scheme are explored in \ref{sec_training}.

\subsection{Prompt-able UNet (PUNet)}
\label{sec_punet}
Instead of using large pre-trained models, as done in the NLP community, we limit ourselves to the investigation of easily applicable small models that can be readily applied to medical data, such as CT and MRT images. In the following, we introduce the dedicated architecture alongside task-dependent prompt tokens. Our prompt tokens can be considered a set of learnable instructions. The tokens aggregate all task-dependent information to achieve a parameter efficient fine-tuning. A new set of prompt tokens is provided for each task with subsets of the tokens representing respective classes for the binary as well as the multi-class case. The frozen backbone model remains task agnostic and thereby class agnostic. Accordingly, the model needs to learn to adapt the embedding predictions and its resulting segmentation masks in dependence of the given prompts. Once learned, the prompt sets can be swapped in accordance to the desired classes (labels) to be predicted. The use of prompts for different training variations is further described in Section~\ref{sec_training}.

We inject the tokens deeply into the network. This allows for an intermediate adaptation of encoded image content throughout the network. Hereby, attention layers provide a structured way to combine and process the heterogeneous encoded image and prompt information. We include memory efficient shifted window (SWin) attention blocks \citep{liu2021swin} in our architecture, which quickly became popular in the medical field \citep{cao2021swin, tang2022self}. Prior works, such as the UNETR \citep{hatamizadeh2022unetr} and SWinUNETR \citep{cao2021swin}, considered a transformer architecture that integrates SWin blocks alongside patch merging and down-projection (linear) layers for dimension reduction. Dense predictions are subsequently retrieved by classical convolutional decoders. The Swin-UNet \cite{cao2021swin} extended the model by integrating SWin blocks throughout an encoder-decoder with intermediate patch merging as well as patch expansion layers. However, we take a step back and employ conventional downstream convolutions (with a stride of 2) in the encoder and rely on linear upsampling layers in the decoder. For effective deep adaptation to a downstream task, we introduce prompting on all resolution levels of this prompt-able UNet (PUNet) architecture. The PUNet architecture is illustrated in Figure \ref{fig:instructed_architecture}. We note that in this work there is no class token, a common constitutent of ViT approaches \citep{dosovitskiy2020image, jia2022visual}, since we perform only segmentation and no classification. To keep a moderate parameter number, a single prompt-able block is applied after each down convolution and before each upsampling layer, replacing most convolutional layers of a traditional UNet. The attention layers operate on patches of $2 \times 2$ for the highest resolution to further save on memory. As such, a convolutional layer with a stride of 2 is being used as first layer. Accordingly, a final upsampling layer is applied to output predictions to achieve the original image (input) resolution. Thus, we are able to provide an architecture that fits the imaging data and allows for the placement of prompt-able blocks throughout the network. A weighted similarity aggregation $\pi_{\mathrm{seg}}$ to a set of prompt tokens precedes the final upsampling to generate class probabilities for the segmentation masks $\mathbf{S} \in \mathbb{R}^{W\times H \times M}$ for $M$ classes (see Section \ref{sec_aggregation}).

\subsubsection{Prompt-able SWin (PSWin) blocks}
\label{sec_pswin}
For this work, we propose to use prompt-able shifted window (PSWin) blocks containing attention layers. Like in previous work, a non-promptable variant of such a block consists of two known transformer blocks, with the content being windowed for the first and spatially shifted and subsequently windowed for the second block. A window size of $8 \times 8$ is used for both transformer blocks and a shift of $4 \times 4$ is applied to the content prior to windowing in the second block. To follow the windowed attention scheme with regular convolutional and upsampling layers in between, the content is rearranged in the respective windowed and shifted windowed content for each promptable block and subsequently reordered in its original embedding shape. The prompt-able block is depicted as part of Figure~\ref{fig:instructed_architecture}.

The windowed content within a transformer block can be jointly processed alongside prompt tokens. As such, the integrated prompt tokens provide learned information about the target objective within each prompt-able block. This means, for the attention layer within a transformer block the prompt tokens are broadcasted and concatenated to each windowed content and processed in a prompt-able multi-head attention (PMA) scheme. We note, that the prompt tokens can be inserted objective-dependent for each batch element. The tokens are passed alongside image encodings to the attention layers which themselves are kept frozen after the pre-training phase. As such, a windowed content $\mathbf{X}^\mathrm{w} \in \mathbb{R}^{N_{\mathrm{w}}\times C}$ with $N_{\mathrm{w}}$ elements and $C$ channels is passed to a prompt-able block together with a set of concatenated prompt tokens $\mathbf{P} = [\mathbf{P}_1, \mathbf{P}_2, ..., \mathbf{P}_M] \in \mathbb{R}^{N_{\mathrm{p}}\times C}$ with $N_{\mathrm{p}}$ total tokens and $M$ subsets of tokens. The concatenation $[\cdot, \cdot]$ is performed along the first dimension. A subset of tokens $\mathbf{P}_m \in \mathbb{R}^{T \times C}$ is comprised of $T$ tokens that represent a specific class $m$. Similarly, $T$ tokens are used per class $m$ in the weighted aggregation scheme introduced in Section \ref{sec_aggregation}. We rely on multiple tokens for each output class to ease learning variable representations for each respective class. For $M$ active classes, this leads to $N_{\mathrm{p}} = M \cdot T$ total tokens that get processed alongside the windowed content in a PMA layer with
\begin{equation} \label{eq_prompt}
    \begin{split}
        \mathbf{X}^{\mathrm{w}, l+1} & = \mathrm{PMA}\left(\left[\mathbf{X}^{\mathrm{w},l}, \mathbf{P}^{l} \right]\right).\\
    \end{split}
\end{equation}
for concatenated inputs $\mathbf{X}^{\mathrm{w}, l}$ and $\mathbf{P}^{l}$ at layer $l$.

Similar forms of this joint processing by an attention layer have been shown to be effective in \citep{lester2021power,jia2022visual}. They represent a simplified form of \citep{li2021prefix} since prompts are concatenated to the respective layer input, instead of using two dedicated prompt sets for queries and keys respectively. Different from the stated related work, our prompt tokens are passed to windowed content. The similarity score within a head $h$ of a PMA layer is calculated as $\mathrm{score}(\mathbf{Q}_h, \mathbf{K}_h) = \mathbf{Q}_h {\mathbf{K}_h}^{\mathrm{T}} / \sqrt{C_{\mathrm{head}}} \in \mathbb{R}^{N_{\mathrm{w}} \times (N_{\mathrm{w}} + N_{\mathrm{p}})}$ with $C_{\mathrm{head}}$ channels per head, queries $\mathbf{Q}_h = \mathbf{W}_h^{\mathrm{q}} {\mathbf{X}^{\mathrm{w},l}}^{\mathrm{T}} \in \mathbb{R}^{N_w\times C_{\mathrm{head}}}$ containing only the windowed content (and no instructions), and keys $\mathbf{K}_h = \mathbf{W}_h^{\mathrm{k}} \left[ \mathbf{X}^{\mathrm{w},l}, \mathbf{P}^{l} \right]^{\mathrm{T}}  \in \mathbb{R}^{(N_w + N_p)\times C_{\mathrm{head}}}$ containing both encoded content as well as prompt tokens. This ensues an attention formulation with head attention $\mathrm{att}_h^{\mathrm{w}}$ and output $\mathbf{X}^{\mathrm{w}, l+1}$ of the next layer $l+1$ with
\begin{equation}
    \begin{split}
    \mathrm{att}_h^{\mathrm{w}} & = \mathrm{softmax}(\mathrm{score}(\mathbf{Q}_h, \mathbf{K}_h)) \mathbf{V}_h \in \mathbb{R}^{N_{\mathrm{w}}\times C_{\mathrm{head}}} \\
    \mathbf{X}^{\mathrm{w},l+1} &= [\mathrm{att}_1^{\mathrm{w}}, ..., \mathrm{att}_{\tilde{H}}^{\mathrm{w}}]{\mathbf{W}^{\mathrm{o}}}^{\mathrm{T}} \in \mathbb{R}^{N_w \times C}
    \end{split}
\end{equation}
for values $\mathbf{V}_h = \mathbf{W}_h^{\mathrm{v}} \left[ \mathbf{X}^{\mathrm{w},l}, \mathbf{P}^{l} \right]^{\mathrm{T}}  \in \mathbb{R}^{(N_w + N_p)\times C_{\mathrm{head}}}$, a projection matrix $\mathbf{W}^{\mathrm{o}} \in \mathbb{R}^{h\cdot C_{\mathrm{head}} \times C}$ and $\tilde{H}$ heads. The resulting full transformer block consists of four sub-layers with residual connections. Starting from windowed rearranged encoded content $\mathbf{X}^{\mathrm{w},l} \in \mathbb{R}^{H' \cdot W' \cdot D' \times C}$ from an encoding $\mathbf{X}^l \in \mathbb{R}^{H \times W \times D \times C}$ at layer $l$, we apply the four steps formulated in equation \ref{eq_swin} to each window. Hereby, an PMA layer $l$ receives a separate set of prompts $\mathbf{P}^{l+n} \in \mathbb{R}^{N_{\mathrm{p}} \times C}$ for the $n$th step. This set is provided to each windowed content $\mathbf{X}^{\mathrm{w},l+n}$ at the respective sub-step.
    \begin{equation} \label{eq_swin}
        \begin{split}
            \mathbf{X}^{\mathrm{w},l+1} & = \mathrm{W-PMA}(\mathrm{IN}(\left[\mathbf{X}^{\mathrm{w},l}, \mathbf{P}^{l}\right])) + \mathbf{X}^{\mathrm{w},l} \\
            \mathbf{X}^{\mathrm{w},l+2} & = \mathrm{Linear}(\mathrm{IN}(\mathbf{X}^{\mathrm{w},l+1})) + \mathbf{X}^{\mathrm{w},l+1} \\
            \mathbf{X}^{\mathrm{w},l+3} & = \mathrm{SW-PMA}(\mathrm{IN}(\left[\mathbf{X}^{\mathrm{w},l+2}, \mathbf{P}^{l+2}\right])) + \mathbf{X}^{\mathrm{w},l+2} \\
            \mathbf{X}^{\mathrm{w},l+4} & = \mathrm{Linear}(\mathrm{IN}(\mathbf{X}^{\mathrm{w},l+3})) + \mathbf{X}^{\mathrm{w},l+3} \\
        \end{split}
    \end{equation}
with W-PMA and SW-PMA representing windowed and shifted windowed PMA blocks as well as instance norms (IN) \cite{ulyanov2016instance} in between the PMA and linear layers. Note, that the typical MLP blocks at the end of an attention block \cite{vaswani2017attention} are reduced to a singular linear layer to keep an overall similar parameter budget to the SwinUNETR.

\subsubsection{Heterogeneous Bias Scores}
\label{sec_bias}

We introduce attention bias scores to the combination of prompt tokens and windowed embedded image content within the joint attention scheme. As these are two heterogeneous inputs processed by the same PMA layer, we account for it by dedicated bias score generation schemes for relevant entries.  For the encoded image content, an additive positional bias $\mathbf{B}^{\mathrm{content}} \in \mathbb{R}^{N_{\mathrm{w}} \times N_{\mathrm{w}}}$ is incorporated in the attention scheme with scores based on relative spatial distances \citep{ramachandran2019stand, cordonnier2019relationship}. This provides spatial locality to the attention layers of the architecture. In comparison, convolutional kernels provide this locality implicitly. In addition, a learnable bias score $\mathbf{B}^{\mathrm{prompt}} \in \mathbb{R}^{N_{\mathrm{w}} \times N_{\mathrm{p}}}$ manipulating the attention score between prompt tokens $\mathbf{P}$ and encoded windowed content $\mathbf{X}_w$ is proposed. This enables us to also optimize the degree to which encoded content at each attention layer is influenced by its prompts. Further bias scores are not required, since the prompts are not processed beyond the attention operation. As such, we generate a head dependent bias matrix $\mathbf{B}_h = [{\mathbf{B}_h^{\mathrm{content}}}, {\mathbf{B}_h^{\mathrm{prompt}}}] \in \mathbb{R}^{N_{\mathrm{w}} \times (N_{\mathrm{w}} + N_{\mathrm{p}})}$ for each head $h$ that can be added to the head dependent attention score values $\mathrm{score}(\mathbf{Q}_h, \mathbf{K}_h)$ of windowed content and prompts. Overall, the resulting biased attention score yields $\mathrm{score}(\mathbf{Q}_h, \mathbf{K}_h) = \mathbf{Q}_h {\mathbf{K}_{h}}^{\mathrm{T}}/ \sqrt{C_{\mathrm{head}}} + \mathbf{B}_h / \sqrt{C_{\mathrm{bias}}}$ with $C_{\mathrm{bias}}$ bias score channels. 

First, we rely on learned bias scores $\mathbf{B}_h^{\mathrm{content}}$ that are added to the attention scores of the windowed content based on relative spatial distances. These scores are dependent on the relative x and y positional pixel distances $d^\mathrm{row}(\mathrm{pos}_i, \mathrm{pos}_j)$, $d^\mathrm{col}(\mathrm{pos}_i, \mathrm{pos}_j)$ for each element $i$ and $j$ within a respective $N_{\mathrm{w}} \times N_{\mathrm{w}}$ window. Biases are calculated for each dimension based on the row and column differences. More specifically their resulting scalar scores are
        \begin{equation} \label{eq_pos_relative}
            \begin{split}
                b_h^{\mathrm{row}}\left(d^\mathrm{row}\right) & = \mathbf{w}_h^{\mathrm{row}} \cdot \mathbf{E}^{\mathrm{row}}\left[ d^\mathrm{row} \right] \in \mathbb{R} \\
                b_h^{\mathrm{col}}\left(d^\mathrm{col}\right) & = \mathbf{w}_h^{\mathrm{col}} \cdot \mathbf{E}^{\mathrm{col}}\left[ d^\mathrm{col} \right] \in \mathbb{R} \\
            \end{split}
        \end{equation}
for the dot product $\cdot$ , a learned weight vector $\mathbf{w}^h \in \mathbb{R}^{C_{\mathrm{bias}}}$ of a head $h$, and a learned embedding for the relative distance $d^{\mathrm{row}}$ in $\mathbf{E}^{\mathrm{row}} \in \mathbb{R}^{N_d \times C_{\mathrm{bias}}}$ as well as the relative distance $d^{\mathrm{col}}$ in $\mathbf{E}^{\mathrm{col}} \in \mathbb{R}^{N_d \times C_{\mathrm{bias}}}$ with $N_d$ containing all possible distances. Scalar scores are aggregated in $\mathbf{B}_h^{\mathrm{content}}$ according to the paired distances and averaged across both dimensions with $\mathbf{B}_h^{\mathrm{content}} = (\mathbf{B}_h^{\mathrm{row}} + \mathbf{B}_h^{\mathrm{col}}) / 2$ before being integrated into the overall bias matrix $\mathbf{B}_h$.

Secondly, we integrate bias scores for attention scores between prompts and windowed encodings. For each prompt token $t$ we calculate a bias entry
        \begin{equation} \label{eq_pos_cross}
            \begin{split}
                b_h^{\mathrm{prompt}}\left(t\right) & = \mathbf{w}_h^{\mathrm{prompt}} \cdot \mathbf{E}^{\mathrm{prompt}}\left[ t \right] \in \mathbb{R} \\
            \end{split}
        \end{equation}
with a head-dependent weight vector $\mathbf{w}_h^{\mathrm{prompt}} \in \mathbb{R}^{C_{\mathrm{bias}}}$, and a learned embedding matrix $\mathbf{E}^{\mathrm{prompt}} \in \mathbb{R}^{N_{\mathrm{p}} \times C_{\mathrm{bias}}}$. The entries of the prompt bias scores $\mathbf{B}_h^{\mathrm{prompt}}$ are unique for each token $t$ of a prompt set $\mathbf{P}$, yet, the same scores are broadcasted to the whole windowed content $\mathbf{X}^{\mathrm{w}}$ (across all windows).

\subsubsection{Cosine Similarity Aggregation}
\label{sec_aggregation}
In addition to the prompt-able blocks, we employ a cosine similarity aggregation in conjunction with learnable prompt tokens to perform token-dependent class predictions. Unlike typical fine-tuning and prior prompting approaches, as explored by \cite{jia2022visual}, there is no requirement for a task-specific linear output layer or entire decoder to project representations into the right amount of output neurons. This allows for greater flexibility, since different segmentation classes as well as different amounts thereof can be predicted in each batch element depending on the selected prompt tokens. Segmentation predictions are generated by means of a cosine similarity calculation between the decoder output $\mathbf{F} \in \mathbb{R}^{W \times H \times C}$ and a final learnable set of prompt tokens $\mathbf{P}^{\mathrm{seg}} \in \mathbb{R}^{N_p \times C}$ (see Figure \ref{fig:instructed_architecture}). The result of this operation are the desired class probabilities $\mathbf{S} \in \mathbb{R}^{W\times H \times M}$ for each available class. This aggregation scheme can be considered an instance of prototypical networks \citep{snell2017prototypical} in which clustered prototypes are replaced by learned prompt tokens. A similar approach has been explored by \citep{strudel2021segmenter} on natural images. However, different from unique class tokens that are processed alongside regular encoded content in their decoder, we employ several learned prompt tokens directly representing respective downstream classes.

A cosine similarity measure is employed between a prompt token $\mathbf{P}_{m, t_m}^{\mathrm{seg}} \in \mathbb{R}^C$ with token index $t_m$ of class $m$ together with an embedding vector $\mathbf{F}_{i,j} \in \mathbb{R}^C$ for an entry with indices $i,j$.
\begin{equation}
\label{eq_sim}
    \tilde{\mathbf{S}}_{i,j,m,t_m} = \mathrm{sim}\left(\mathbf{F}_{i,j}, \mathbf{P}_{m, t_m}^{\mathrm{seg}}\right) = \frac{\mathbf{F}_{i,j} \cdot \mathbf{P}_{m, t_m}^{\mathrm{seg}}}{\lVert \mathbf{F}_{i,j} \rVert \: \lVert \mathbf{P}_{m, t_m}^{\mathrm{seg}} \rVert} \in \mathbb{R}
\end{equation}
Instead of an average of all tokens $T$ for each class $m$, a weighted similarity score is used to retrieve class probabilities $\mathbf{S}$ with entries
\begin{equation} \label{eq_aggregation}
    \mathbf{S}_{i,j,m} = \sum_{t_m=1}^{T} \mathrm{softmax}(\tilde{\mathbf{S}}_{i,j,m} / \tau_{\mathrm{agg}})_{t_m} \odot \tilde{\mathbf{S}}_{i,j,m,t_m} \in \mathbb{R}
\end{equation}
for the element-wise product $\odot$ and a temperature parameter $\tau_{\mathrm{agg}}$. The softmaxed weight does not influence the backward pass during training, since it is excluded from gradient updates. This way not every prompt token has to align to all relevant image content representations belonging to the same class.

\subsection{Dense Self-Supervision}
\label{sec_self}

To establish robust anatomical representations suitable for further downstream tasks on medical imaging data, we employ a combination of data augmentation and a dedicated self-supervision scheme of aligning online generated prototypes between a teacher network and student networks. The whole scheme is depicted  in Figure \ref{fig:training_scheme}. It generates embeddings where anatomically similar regions are represented close to each other. It incorporates a momentum model with an EMA updated teacher and students, prominently used in \citep{grill2020bootstrap}. We incorporate two students, one processing a smaller input $\mathbf{\hat{I}'} \in \mathbb{R}^{W^{\mathrm{s}_1}\times H^{\mathrm{s}_1}}$ than the teacher input $\mathbf{\tilde{I}} \in \mathbb{R}^{W^{\mathrm{t}}\times H^{\mathrm{t}}}$ and the second one $\mathbf{\hat{I}''} \in \mathbb{R}^{W^{\mathrm{s}_2}\times H^{\mathrm{s}_2}}$ more severely cropped to enforce robust embeddings $\mathbf{F}$ with focus on global as well as more local context alike. The smaller student FOVs are cropped out of the teacher FOV. Both students share the same underlying network weights. This multi-crop strategy has been proven beneficial in approaches proposed like \citep{chen2020simple, caron2020unsupervised}. Furthermore, the proposed student teacher combination naturally allows for the integration of partially masked content in each student view as part of the augmentation pipeline. Thus, embeddings have to be robustly learned to ensure predicting an assignment similar to the one provided by a non-masked teacher, further enforcing the estimation of output representations based on context.  Different from most approaches mentioned in \ref{sec_refs_self_sup}, we are solely interested in predicting a learned embedding, despite adverse effects being present in the input views, instead of reconstructing the original input itself.

\subsubsection{Contrastive Prototype Assignments (CPA)}
\label{sec_cpa}

\begin{figure*}[h]
    \centering
    \includegraphics[width=\textwidth]{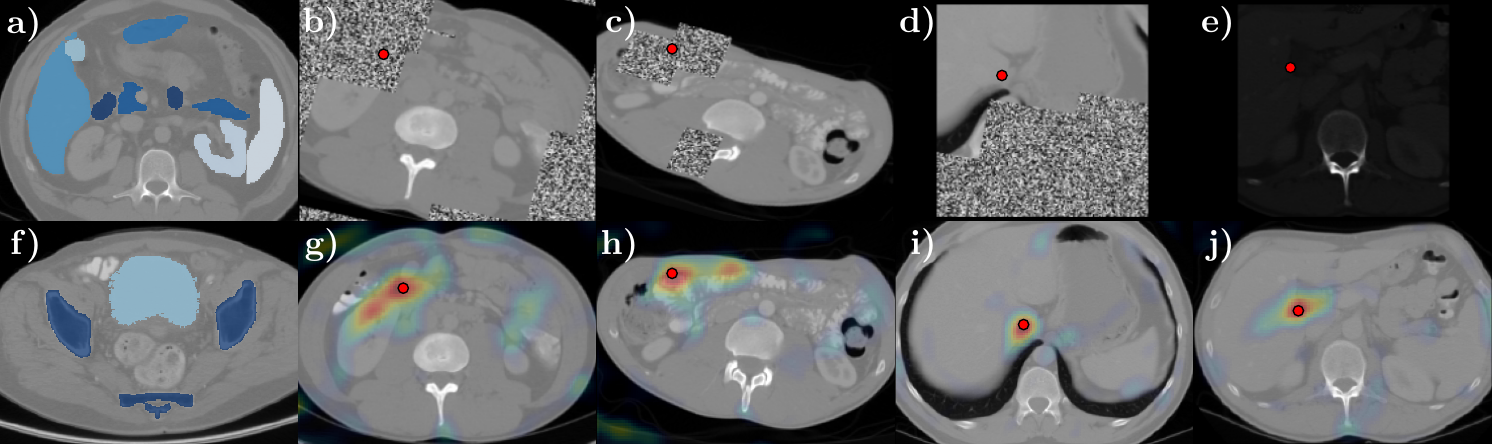}
    \caption{a/f) Exemplary slices of the TCIA/BTCV and CT-ORG dataset, with annotated masks shown in shades of blue, b-e) augmented student views with masked regions or strong contrast adjustments, g-j) respective teacher views with overlays of the cosine similarity of the predicted teacher embedding $\mathbf{F}^{\mathrm{t}}$ and the student embedding $\mathbf{F}^{\mathrm{s}}_{i,j}$ at an arbitrary selected point of interest (indicated by a red dot) with indices $i,j$ in the corresponding student view. Highly similar regions appear red in the teacher view. The approach learns a robust embedding that enforces context learning, and is thus able to generate proper similarities, despite the origin region being severely affected. Note, that teacher augmentations have been disabled for better visual clarity in this illustration.}
    \label{fig_data}
\end{figure*}

To facilitate the learning of an effective embedding $\mathbf{F}$, the two students are penalized to predict the same contrastive prototype assignments (CPA) as its teacher. For each student teacher combination a cross entropy loss $\mathcal{L}_\mathrm{CE}$ is applied.
\begin{equation}
    \mathcal{L}_\mathrm{CPA} = \frac{1}{2}\sum_{n=1}^2 \mathcal{L}_{\mathrm{CE}}\left(\mathbf{\Phi}^{\mathrm{s}_n}, \mathbf{\Phi}^{\mathrm{t}}\right)
\end{equation}
The teacher provides a guiding assignment $\mathbf{\Phi}^{\mathrm{t}} \in \mathbb{R}^{W^{\mathrm{t}} \times H \times C}$ which has to be replicated by the student by its assignment $\mathbf{\Phi}^{\mathrm{s}_n} \in \mathbb{R}^{W^{\mathrm{s}_n} \times H \times C}$. Thus, we encourage the prediction of similar output embeddings $\mathbf{F}^{\mathrm{t}}$ of the teacher and $\mathbf{F}^{\mathrm{s}}$ of a student by enforcing identical assignments to online generated prototypes $\mathbf{C}^{\mathrm{p}} \in \mathbb{R}^{N_k \times C}$ for $N_k$ prototypes. We calculate similarity assignment entries via
\begin{equation}
\label{eq_assignment}
\begin{split}
    \mathbf{\Phi}_{i,j,k} = \pi_{\mathrm{sim}}(\mathbf{F}, \mathbf{C}^{\mathrm{p}})_{i,j,k} & = \mathrm{softmax}(\mathrm{sim}(\mathbf{F}, \mathbf{C}^{\mathrm{p}})/\tau_{\mathrm{assign}})_{i,j,k} \\
    & = \frac{\exp\left(\mathrm{sim}(\mathbf{F}_{i,j}, \mathbf{c}_k / \tau_{\mathrm{assign}}\right)}{\sum_{\tilde{k = 1}}^{N_k} \exp\left(\mathrm{sim}(\mathbf{F}_{i,j}, \mathbf{c}_{\tilde{k}}^{\mathrm{p}}) / \tau_{\mathrm{assign}}\right)}
\end{split}
\end{equation}
for a pixel with indices $i,j$ and a prototype $k$. We rely on the cosine similarity operator $\mathrm{sim}$ defined in equation \ref{eq_sim} and a temperature $\tau_{\mathrm{assign}}$. For the teacher assignment a smaller temperature value is applied in the softmax than is used in the student, encouraging progressively confident predictions to the generated clusters \citep{assran2021semi}. To get the proper target for a student pixel at $\mathrm{pos}^{\mathrm{s}}_{i,j}$, the spatially closest teacher assignment $\mathbf{\Phi}^{\mathrm{t}}_{u,v}$ at position $\mathrm{pos}^{\mathrm{t}}_{u,v}$ is queried. This is possible, since the position $\mathrm{pos}_{i,j}$ of an output pixel with indices $i,j$ is known by a common underlying coordinate grid of the original 3D volume, which is shifted in accordance to spatial image augmentations. For computational efficiency $\mathbf{F}^{\mathrm{t}}$ and $\mathbf{F}^{\mathrm{s}}$ are sampled with a factor of 2 for the loss calculation. Hereby, a random spatial jitter (shift) is applied before sampling $\mathbf{F}^\mathrm{s}$ to prevent potential gridding artefacts.

We note, that this CPA loss $\mathcal{L}_\mathrm{CPA}$ follows the contrastive formulation known from \citep{oord2018representation}. However, since the loss target $\mathbf{\Phi}^{\mathrm{t}}$ is an online generated soft assignment, our self-supervision scheme does not enforce a strict division into positives (that should be attracted) and negatives (that should be repelled). This allows for the concurrent application of a segmentation supervision (see Section \ref{sec_training}), where the predicted embedding has to adhere to multiple objectives at the same time. The implicit separation also differentiates our method from approaches, such as the self-supervised anatomical embedding of \cite{chaitanya2020contrastive, yan2022sam}, where positives and negatives are sampled based on their proximity to the anchor location for multiple (global and local) embeddings. The included pre-defined separation heuristics may be in violation to a second (segmentation) loss target.

\subsubsection{Online Prototype Generation}
\label{sec_prototypes}

\begin{figure*}[h]
    \centering
    \includegraphics[width=\textwidth]{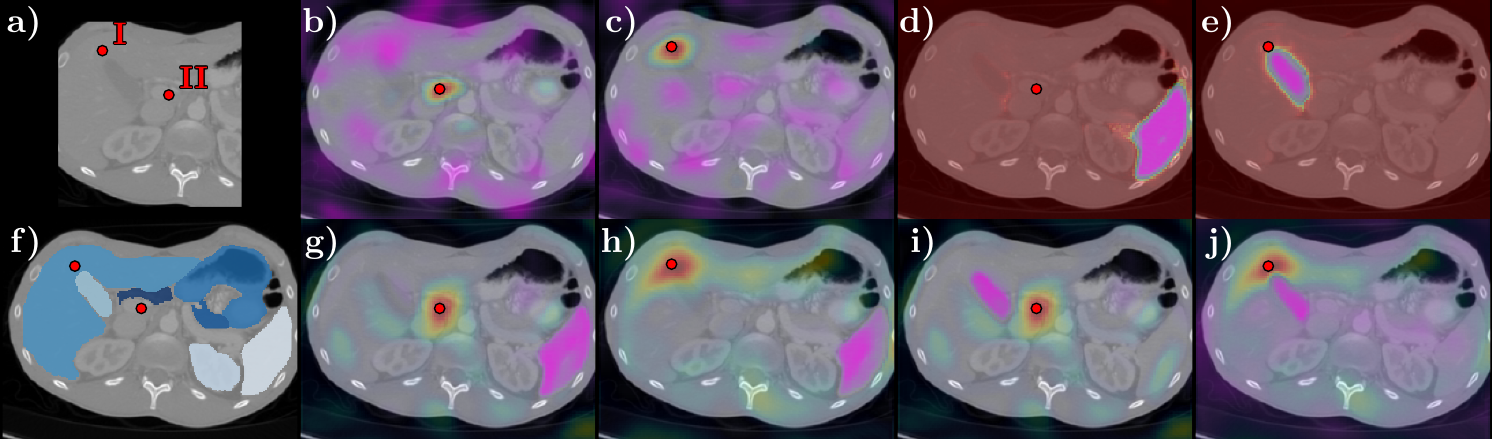}
    \caption{Visualization of cosine similarities between predicted teacher embeddings $\mathbf{F}^{\mathrm{t}}$ and a student embedding $\mathbf{F}^{\mathrm{s}}_{i,j}$ at arbitrary selected points of interest (red dots with labels I and II) with respective indices $i,j$ for the student view shown in (a). f) The respective teacher slice with points of interest located closely above the gall bladder (I) and below the pancreas (II). b-c) Teacher views with cosine similarities for the self-supervised pre-trained model for respective points of interest. The similarity is densely concentrated around the queried point. d-e) Cosine similarities of the segmentation supervised pre-trained model with prompt tokens active for the spleen (d) and the gallbladder (e). Pink regions indicate highly dissimilar regions (inverse of the similarity map) and serve as indication of the generated segmentation masks. There is no difference with respect to location outside of the active target region. g-j) Similarity maps for the combined self- and segmentation supervised pre-trained model. Densely concentrated similarities are visible alongside highly dissimilar active target regions. In (j) it can be seen, that the prompt tokens can successfully alter regions that showed high similarity in (h) but belong now to the active target region (by the change from red (in h) to pink (in j) in the periphery of the point of interest.}
    \label{fig_dimensions}
\end{figure*}

For our case, bootstrapping codes \cite{caron2020unsupervised} or maintaining a relevant queue \cite{gidaris2021obow} is non-trivial given the partial FOV of the input data and the textural and shape variabilities as well as inhomogeneities encountered in medical images. For similar reasons, we also do not rely on codes represented by superpixels, such as by aggregated online generated \citep{henaff2022object}. For medical images, superpixels require a deliberate offline scheme \citep{ouyang2020self} suitable for the underlying data. Instead, our approach falls in the category of derived prototypes. We rely on an iterative clustering scheme based on calculated assignments to update prototype clusters. Hereby, our generation process can be considered as a k-means clustering with spatially weighted soft assignments. The online generated prototypes $\mathbf{C}^{\mathrm{p}}$ represent clustered representatives of the teacher embedding $\mathbf{F}^{\mathrm{t}}$. For simplicity, we reuse the similarity operator $\pi_{\mathrm{sim}}$ to assign features to prototypes. As such, we calculate prototype assignments $\mathbf{\Phi}^{\mathrm{p}} \in \mathbb{R}^{W^{\mathrm{t}} \times H \times C} = \pi_{\mathrm{sim}}(\mathbf{F}^{\mathrm{t}}, \mathbf{C}^{\mathrm{p}})$ similar to equation \ref{eq_assignment}. Initial prototypes are drawn from interpolated entries of the predicted teacher embedding $\mathbf{F}^{\mathrm{t}}$ at selected seed points. The seed points are defined by a grid overlayed on the teacher embeddings. The grid size and as such the amount of clusters is being determined by a reduction factor with respect to the output size of $\mathbf{F}^{\mathrm{t}}$.

To further guide the clustering process, the assignments of each entry to the current prototypes are weighted by a positional bias with $\tilde{\mathbf{\Phi}}^{\mathrm{p}} = \mathbf{W}^{\mathrm{p}} \odot \mathbf{\Phi}^{\mathrm{p}}$. This promotes a spatial proximity to clusters with centroids in a close neighborhood being preferred. The weight vector $\mathbf{W}^{\mathrm{p}}$ is derived from the relative spatial distances between the pixel with indices $i,j$ and the position of the prototype $k$. The weight entries are defined by the relative Gaussian distance
\begin{equation}
    \mathbf{W}^{\mathrm{p}}_{i, j, k} = \exp\left(- \frac{\Vert\mathrm{pos}_{i,j}^{\mathrm{t}} - \mathrm{pos}_k^{\mathrm{p}}\Vert^2}{2 \cdot \sigma^2}\right)
\end{equation}
with $\sigma^2$ being determined based on a pre-defined full width at half maximum (FWHM) parameter with the width given in pixels. For each iteration a new resulting prototype centroid $\mathbf{C}_k^{\mathrm{p}}$ and its new position $\mathrm{pos}_{k}$ are derived, based on the degree to which the teacher embedding $\mathbf{F}^{\mathrm{t}}$ belongs to the prior iteration of prototype centroids, represented by the  $\tilde{\mathbf{\Phi}}^{\mathrm{p}}$.
\begin{equation}
    \mathbf{C}_k^{\mathrm{p}} = \frac{\sum_{i,j} \tilde{\mathbf{\Phi}}_{i,j,k}^{\mathrm{p}} \mathbf{F}_{i,j}^{\mathrm{t}}}{\sum_{i,j} \tilde{\mathbf{\Phi}}_{i,j,k}^{\mathrm{p}}} \,,\:\: \mathrm{pos}_{k}^{\mathrm{p}} = \frac{\sum_{i,j} \tilde{\mathbf{\Phi}}_{i,j,k}^{\mathrm{p}} \mathrm{pos}_{i,j}^{\mathrm{t}}}{\sum_{i,j} \tilde{\mathbf{\Phi}}_{i,j,k}^{\mathrm{p}}}
\end{equation}
In essence, we sample features of a momentum-updated teacher and calculate weighted assignment to generate spatially guided online prototype centroids. In turn, these assignments to the newly generated prototypes of the predicted teacher and student embeddings are enforced to be similar by the CPA loss $\mathcal{L}_\mathrm{CPA}$. Working with online generated prototypes, we also avoid the need for further losses, such as prevalent in mutual information maximization schemes \citep{peng2021boosting}, to enforce well distributed assignments. In addition, relying on the online prototype generation brings the benefit of avoiding a complex queue all together without sacrificing convergence to similar representations for similar regions. Contrary, using a queue would require to establish and update a diverse and representative set that represents meaningful targets. Overall, no further guidance such as pre-defined heuristics \citep{chaitanya2020contrastive} or auxiliary labels \citep{ouyang2020self} are required. The scheme shares some similarity with the work of \cite{henaff2022object}, which relies on masks generated by a k-means algorithm which are mapped on to the student FOVs and used as assignment targets. Contrary to \citep{henaff2022object}, we do not use aggregated object representations, but simply use the generated prototypes. Instead, our soft assignment allows for the assignment of pixels to multiple online generated prototypes which may share similar characteristics.

\begin{table*}[t]
\setlength{\tabcolsep}{4pt} 
\centering
\small
\begin{tabular}{ l r c c c c c c c }
 \hline
 \multirow{2}{*}{Method} & \multirow{2}{*}{Parameters} & \multirow{2}{*}{Amount} & \multicolumn{2}{c}{TCIA/BTCV} & & \multicolumn{2}{c}{CT-ORG} \\
 \cline{4-5} \cline{7-8}
 & & & P1 & P2 & & P1 & P2 & \\
 \hline
 nnUNet & $30.0\mathrm{M}$ & \multirow{7}{*}{$100\%$} & $\mathbf{83.94 \pm \phantom{0}4.64 \:\: (84.70)}$ & $-$ & & $\mathbf{91.44 \pm \phantom{0}6.15 \:\: (93.15)}$ & $-$ \\
 UNet & 5.0$\mathrm{M}$ & & $82.45 \pm \phantom{0}4.48 \:\: (82.97)$ & $82.93 \pm \phantom{0}4.11 \:\: (84.39)$ & & $89.01 \pm \phantom{0}6.95 \:\: (90.53)$ & $89.90 \pm \phantom{0}5.75 \:\: (91.56)$ \\ 
 UNETR & 87.3$\mathrm{M}$ & & $78.31 \pm \phantom{0}5.64 \:\: (79.44)$ & $79.50 \pm \phantom{0}4.80 \:\: (81.14)$ & & $87.96 \pm \phantom{0}6.64 \:\: (90.24)$ & $88.83 \pm \phantom{0}6.06 \:\: (90.35)$ \\ 
 SwinUNETR & 6.3$\mathrm{M}$ & & $82.15 \pm \phantom{0}4.41 \:\: (83.37)$ & $82.18 \pm \phantom{0}3.96 \:\: (82.21)$ & & $89.67 \pm \phantom{0}6.26 \:\: (91.75)$ & $\mathbf{90.18 \pm \phantom{0}5.98 \:\: (92.05)}$ \\ 
 PUNet (binary) & 6.8$\mathrm{M}$ & & $82.07 \pm \phantom{0}4.38 \:\: (81.85)$ & $\mathbf{83.45 \pm \phantom{0}3.64 \:\: (84.02)}$ & & $88.50 \pm \phantom{0}6.48 \:\: (90.13)$ & $89.74 \pm \phantom{0}6.13 \:\: (91.39)$\\
 PUNet (multi c.) & 7.3$\mathrm{M}$ & & $79.92 \pm \phantom{0}5.52 \:\: (79.66)$ & $80.98 \pm \phantom{0}5,17 \:\: (81.03)$ & & $90.64 \pm \phantom{0}6.01 \:\: (92.45)$ & $89.75 \pm \phantom{0}6.74 \:\: (91.28)$ \\ 
 PUNet (fixed) & 6.8$\mathrm{M}$ & & $82.50 \pm \phantom{0}3.88 \:\: (82.15)$ & $82.45 \pm \phantom{0}3.70 \:\: (82.85)$ & & $90.25 \pm \phantom{0}6.09 \:\: (91.95)$ & $89.70 \pm \phantom{0}6.12 \:\: (91.02)$ \\ 
 \hline
 nnUNet & $30.0\mathrm{M}$ & \multirow{7}{*}{$10\%$} & $66.64 \pm 10.92 \:\: (67.64)$ & $-$ & & $\mathbf{88.06 \pm 10.48 \:\: (91.50)}$ & $-$ \\
 UNet & $5.0\mathrm{M}$ & & $68.87 \pm \phantom{0}6.85 \:\: (68.63)$ & $72.13 \pm \phantom{0}6.11 \:\: (72.66)$ & & $86.05 \pm 10.79 \:\: (89.51)$ & $85.90 \pm 12.04 \:\: (90.10)$ \\ 
 UNETR & $87.3\mathrm{M}$ & & $65.85 \pm \phantom{0}7.88 \:\: (67.14)$ & $70.34 \pm \phantom{0}5.42 \:\: (70.61)$ & & $84.68 \pm 11.25 \:\: (88.68)$ & $85.82 \pm \phantom{0}9.43 \:\: (88.22)$ \\ 
 SwinUNETR & $6.3\mathrm{M}$ & & $67.76 \pm \phantom{0}7.20 \:\: (66.99)$ & $\mathbf{72.71 \pm \phantom{0}5.74 \:\: (72.42)}$ & & $86.00 \pm 11.03 \:\: (89.80)$ & $86.49 \pm 11.06 \:\: (89.52)$ \\ 
 PUNet (binary) & $6.8\mathrm{M}$ & & $70.78 \pm \phantom{0}7.04 \:\: (72.41)$ & $72.03 \pm \phantom{0}7.14 \:\: (72.60)$ & & $86.39 \pm 13.10 \:\: (90.71)$ & $\mathbf{87.97 \pm \phantom{0}9.10 \:\: (91.01)}$ \\
 PUNet (multi c.) & $7.3\mathrm{M}$ & & $67.59 \pm \phantom{0}7.30 \:\: (68.76)$ & $69.91 \pm \phantom{0}7.07 \:\: (70.50)$ & & $86.66 \pm \phantom{0}9.64 \:\: (89.68)$ & $86.72 \pm 10.97 \:\: (90.82)$ \\ 
 PUNet (fixed) & $6.8\mathrm{M}$ & & $\mathbf{71.00 \pm \phantom{0}7.80 \:\: (72.58)}$ & $71.43 \pm \phantom{0}7.35 \:\: (72.32)$ & & $86.83 \pm 10.53 \:\: (90.18)$ & $87.30 \pm \phantom{0}8.82 \:\: (89.84)$ \\ 
 \hline
\end{tabular}
\caption{Mean $\pm$ standard deviation (median) of the segmentation performance expressed by the Dice Similarity Coefficient (DSC, in \%) for the two investigated TCIA/BTCV and CT-ORG datasets. The nnUNet is trained within its framework. All other UNet baseline variants (UNet, UNETR, SwinUNETR) are integrated by means of the MONAI framework and trained in identical fashion to the prompt-able UNet (PUNet) variants. The variants differ by using prompt tokens for binary predictions (binary), concurrent prompt tokens for each class (multi class) and a fixed output layer (fixed) in which case no prompt tokens are passed to the prompt-able blocks. Architectures are investigated for a single phase 1 (P1) training where only a segmentation loss is applied for the present case and a two phase performance (P2) where a pre-training is applied prior to a full fine-tuning on the respective classes. In addition, runs are performed for $100\%$ and $10\%$ of all available training data.}
\label{table_baselines}
\end{table*}

\subsection{Training Procedure}
\label{sec_training}
With the prompt-able architecture and the dedicated self-supervision scheme in place, we can train the given neural network and subsequently adapt the model in a parameter-efficient manner to an unseen downstream task. For clarity, we introduce two distinct training stages: phase 1 (P1) where a fully learnable model is (pre-)trained and phase 2 (P2) where the pre-trained model is adapted by selective parameter training to the downstream tasks. We investigate the performance of the proposed approach by considering several experiments, where the model is frozen in P2. In some cases, we are also interested in full fine-tuning of the whole model and consider this as non-frozen adaptation. When a model is denoted as frozen, the bulk of the architecture is non-trainable, but prompts $\mathbf{P}$ throughout the network, including $\mathbf{P}^{\mathrm{seg}}$, as well as attention bias scores $b_h^{\mathrm{prompt}}$ remain trainable.

In phase 1, the self-supervision loss $\mathcal{L}_\mathrm{CPA}$ is applied. If this remains the sole loss, the prompt-able blocks process only encoded image content, with the prompts and its bias scores not being inserted in the PMA layer. However, the prompts are considered if a concurrent segmentation loss is applied, making use of the prompts and thereby enabling the network to learn to prompt during pre-training. For this case, softmaxed predictions of each pixel $\hat{y}_i$ are penalized by a class-weighted focal loss \cite{lin2017focal}
\begin{equation}
    \mathcal{L}_{\mathrm{focal}} = - \frac{1}{W \cdot H}\sum_{i = 1}^{W \cdot H} \sum_{l = 1}^{M} \alpha_m \left(1 - \hat{y}_i \right)^{\gamma} \log\left(\hat{y}_i\right) y_{i, l}
\end{equation}
with a focusing parameter $\gamma$, one-hot encoded true mask values $y_{i, m}$ and weighting factors $\alpha_m$ for $M$ label categories (classes). $\alpha_m$ is calculated for each class based on a heuristic which builds on the average foreground to background ratios of a training set.

Since the architecture allows for batch element dependent prompt insertion, classes declared to be used for training can be randomly drawn and applied in a joint batch. As such, respective (learned) prompt tokens $\mathbf{P}$ are stored for each combination of classes and drawn in accordance to the selected task. At the same time, the shared neural network weights get trained with respect to all available tasks and their classes. This allows for efficient pre-training not only of a suitable representation induced by the self-supervision, but also class-dependent predictions which aid in enforcing good generalizations to further unseen classes during adaptation. Thus, we keep gradual differences for anatomical regions while learning to adapt the predicted embedding to follow clear separations between known prompt-dependent regions. Hence, in its most advanced form, we train the model on a subset of all available classes and perform the adaptation on a disjoint set. As such, our scheme can be seen as a combination of supervised and self-supervised learning to provide more robust classification or segmentation results in the case of scarce annotations. In phase 2, we rely solely on the aforementioned focal loss $\mathcal{L}_{\mathrm{focal}}$ to penalize the correct prediction of classes based on aggregated class probabilities.

Several variations of the prompting scheme can be considered independent of the training phases. The architecture can be used to provide binary predictions, with each class having their own background and foreground tokens, or with a dedicated set of background and foreground prompt tokens for each class. The latter one describes the typical multi-class case. It is also possible to provide multiple binary segmentations and recombine them in a multi-class segmentation in a post-processing step. Following the typical fine-tuning scheme, one can ignore the instructions and use the proposed network architecture with a fixed linear layer to project embeddings to the desired class probabilities. In that case, a simple linear (projection) layer provides a fixed number of classes instead of using the proposed similarity comparison with prompt tokens.

\section{Results}

\subsection{Experimental Setup}
We conduct our experiments on two publicly available medical CT datasets. The first is a joint set, the TCIA/BTCV dataset \cite{gibson2018automatic}, comprised of $89$ subjects with densely annotated 3D volumes of the The Cancer Imaging Archive (TCIA) Pancreas-CT dataset \cite{roth2016data} and the Beyond the Cranial Vault (BTCV) abdomen dataset \cite{landman2015miccai}. We include eight organ masks, that are part of both sub-groups, namely the spleen, left kidney, gallbladder, liver, esophagus, stomach, pancreas and duodenum.  The second, the CT-ORG dataset \cite{rister2019ct} includes a diverse set of five organs for $139$ subjects. Here the liver, the bladder, both kidneys, but also visually distinct structures such as the lungs and the general bone structure are annotated. This dataset includes contrast and non-contrast enhanced CT scans. We use a fixed random split for all datasets, corresponding to a 70\% training, 10\% validation, and 20\% test split. This results in splits of 60/10/19 subjects for TCIA/BTCV and 97/14/28 subjects for CT-ORG. Exemplary axial slices with overlayed segmentation masks are shown in Figure \ref{fig_data} (a/f). For the joint application of losses, we split the classes of the datasets into two disjoint groups. For TCIA/BTCV, we have the spleen, left kidney, gallbladder, and liver in the first group and the esophagus, stomach, pancreas and duodenum in the second group. For CT-Org the liver, bladder, and kidneys are grouped together with the lungs and bones comprising the second group.

\begin{table*}[t]
\setlength{\tabcolsep}{4pt} 
\centering
\small
\begin{tabular}{ c c l c c c c c c c c }
 \hline
 & & \multirow{2}{*}{Variant} & \multicolumn{2}{c}{P1} & & \multicolumn{2}{c}{TCIA/BTCV} & & \multicolumn{2}{c}{CT-ORG} \\
 \cline{4-5} \cline{7-8} \cline{10-11}
 & & & Seg. & Self & & DSC & ASSD & & DSC & ASSD \\
 \hline
 \multirow{8}{*}{P2} & \multirow{4}{*}{Non-frozen} & Joint & \checkmark & \checkmark & & $83.13 \pm 3.86 \:\: (83.96)$ & $1.92 \pm 0.92 \:\: (1.73)$ & & $87.18 \pm 6.53 \:\: (89.91)$ & $\phantom{0}3.64 \pm \phantom{0}2.83 \:\: (2.97)$ \\ 
 & & Seg. & \checkmark & - & & $83.10 \pm 3.79 \:\: (83.47)$ & $2.02 \pm 1.09 \:\: (1.62)$ & & $\mathbf{89.90 \pm 6.02 \:\: (91.45)}$ & $\mathbf{\phantom{0}3.17 \pm \phantom{0}3.20 \:\: (1.89)}$\\
 & & Self & - & \checkmark & & $\mathbf{83.45 \pm 3.64 \:\: (84.02)}$ & $\mathbf{1.83 \pm 0.92 \:\: (1.55)}$ & & $89.74 \pm 6.13 \:\: (91.39)$ & $\phantom{0}3.29 \pm \phantom{0}3.44 \:\: (2.10)$ \\
 & & Random & - & - & & $82.21 \pm 3.63 \:\: (81.69)$ & $2.05 \pm 1.16 \:\: (1.69)$& & $85.02 \pm 7.28 \:\: (87.80)$ & $\phantom{0}6.80 \pm \phantom{0}7.01 \:\: (4.47)$\\ 
 \cline{2-11}
 & \multirow{4}{*}{Frozen} & Joint & \checkmark & \checkmark & & $\mathbf{79.62 \pm 3.81} \:\: (79.55)$ & $\mathbf{2.30 \pm 1.14 \:\: (1.97)}$ & & $\mathbf{87.23 \pm 6.87 \:\: (88.42)}$ & $\phantom{0}6.59 \pm 12.93 \:\: (2.68)$\\ 
 & & Seg. & \checkmark & - & & $75.73 \pm 5.15 \:\: (75.49)$ & $3.31 \pm 1.76 \:\: (2.61)$ & & $82,13 \pm 7.55 \:\: (83.83)$ & $\phantom{0}8.92 \pm \phantom{0}8.50 \:\: (4.77)$\\
 & & Self & - & \checkmark & & $73.88 \pm 4.69 \:\: (74.82)$ & $3.01 \pm 1.85 \:\: (2.52)$ & & $84.14 \pm 6.88 \:\: (86.51)$ & $\mathbf{\phantom{0}4.68 \pm \phantom{0}4.09 \:\: (3.76)}$\\
 & & Random & - & - & & $60.96 \pm 7.02 \:\: (62.63)$ & $5.35 \pm 2.36 \:\: (4.52)$ & & $73.40 \pm 8.30 \:\: (73.95)$ & $10.11 \pm \phantom{0}6.94 \:\: (8.50)$ \\ 
 \hline
\end{tabular}
\caption{Mean $\pm$ standard deviation (median) of the downstream segmentation performance depicted by the Dice Similarity Coefficient (DSC, in \%) and Average Symmetric Surface Distance (ASSD, in mm) for different training scheme variations on the TCIA/BTCV and CT-ORG datasets. We included the following combinations of self- and segmentation supervision in the pre-training phase 1 (P1): self- and segmentation (joint), segmentation (seg.), self-supervision (self) and a random initialization (random) of the weights without any applied losses in P1. For the downstream phase 2 (P2), we differentiate between a trainable backbone architecture (non-frozen) and a non-trainable architecture (frozen) except for the prompt tokens (and its dedicated attention bias scores). For variants including a segmentation loss, two models are trained with subsets of the classes seen during P1 and a disjoint set seen during P2.}
\label{table_training_schemes}
\end{table*}

In a pre-processing step, all volumes are re-scaled to a target resolution of $1.25\,\mathrm{mm} \times 1.25 \,\mathrm{mm} \times 2.5 \,\mathrm{mm}$ and cropped to an in-plane matrix size of $280 \times 280$ pixels. We set input FOVs to $256\times256$ pixels for the teacher, and $224\times224$ as well as $160\times 160$ pixels for the two students. Facilitated by the MONAI framework \citep{monai2020}, the data augmentation includes a clipped intensity re-scaling, random spatial crops in accordance with the input sizes of the teacher and both students, random bias field, contrast adjustments, histogram and intensity shift and scales as well as affine transformations including rotations, scale and shear changes. In addition, random regions of student inputs are locally masked with dropouts and shuffling of the respective content.

The training procedure differs between the two phases P1 and P2. Phase 1 follows a training process with an Adam optimizer with decoupled weight decay (AdamW) \citep{loshchilov2017decoupled} with a learning rate of $10^{-4}$ for neural network parameters, $10^{-3}$ for prompting parameters, and a weight decay of $10^{-2}$. P1 is applied for 400 epochs of 5000 samples drawn out of the training set per epoch. In Phase 2 an Adam optimizer is used with a learning rate of $5\times10^{-4}$ for eligible network parameters (relevant for ablation architectures) and $5\times 10^{-3}$ for prompt parameters. This phase is shortened to 100 epochs (for computational reasons) and provided with a one cycle learning rate scheduler \citep{smith2019super}. Further, hyperparameters are set empirically. Loss weights of $1.0$ for the supervised loss $\mathcal{L}_{\mathrm{focal}}$ and $10^{-2}$ for the self-supervised loss $\mathcal{L}_\mathrm{CPA}$ are used. A $\gamma$ of $4.0$ is selected for the focal loss. For the self-supervision, a FWHM of $128$ (pixels), a prototype cluster reduction factor of $8$, softmax temperatures $\tau_{\mathrm{assign}}$ of $0.033$ for the teacher and $0.066$ for students are applied. The softmax temperature $\tau_{\mathrm{agg}}$ of the similarity aggregation is set to $0.1$.

Throughout the architecture, batch normalization for convolutions and instance normalization within prompt blocks are used alongside leaky ReLU activations. A depth of 5 levels with 32, 64, 128, 256, 384 hidden channels $C$ for each level is chosen. $\tilde{H} = 8$ attention heads are used for each PMA layer together. Within a PMA layer, we use $C_{\mathrm{head}} = C / \tilde{H}$ channels per head and $C_{\mathrm{bias}} = 32$ channels for instruction bias scores. In addition, $T=16$ prompt token vectors are used to represent a class. 

Predictions are evaluated by means of the Dice similarity coefficient (DSC), as a measure of the overlap quality between ground truth annotation and prediction, and the average symmetric surface distance (ASSD), to assess surface distances between ground truths and predictions. Metrics are calculated from whole 3D volumes. Metrics are reported as mean $\pm$ standard deviation across all test subjects of the average value of all foreground classes if not stated otherwise. In addition median values (in brackets) are mentioned. For hypothesis testing, p-values of a two-sided paired t-test are reported to assess differences in mean populations. A significance threshold of $0.05$ is set for comparisons with respect to our proposed method.

\subsection{Experiments}
In the following, we provide six experiments alongside qualitative examples to show the effectiveness of the aspects proposed in this work. We cover comparisons with state-of-the-art architectures, the impact of the self-supervision scheme, different pre-training combinations, various downstream adaptation approaches besides learning prompt tokens, the insertion positions of prompt tokens, and the behaviour in the annotation scarce case.

Qualitative results of the proposed self-supervised training scheme are portrayed in Figure \ref{fig_data} and \ref{fig_dimensions}. Hereby, we calculate cosine similarities between a predicted teacher embedding $\mathbf{F}^{\mathrm{t}}$ and exemplary points of interest (indicated by a red dot) of the student embedding $\mathbf{F}^{\mathrm{s}}_{i,j}$ for indices $i,j$. The overlay of the resulting similarity map on the teacher FOV depicts highly similar regions in red. For Figure \ref{fig_data}, the similarity of arbitrary selected points in several augmented student views (b-e) is visualized for their corresponding unmasked teacher views (g-j). Despite severe augmentations, including extensive masking, the resulting cosine similarity values, overlayed in the teacher view, enable the identification of the original region. In Figure \ref{fig_dimensions} the cosine similarity for two arbitrary points of interest of a student view (a) and the respective cosine similarity to predicted teacher embeddings are overlayed for the different training scheme variations. In dependence of the application of the losses in the pre-training phase 1, the network is able to identify similar regions for the self-supervision (\textit{self}, b-c), segment annotated regions for the segmentation supervision (\textit{seg.}, d-e) and localize regions while distinctively separating semantic regions for the application of both losses (\textit{joint}, g-j).

We compare our introduced PUNet architecture in conjunction with and without the proposed self-supervision scheme. We provide several popular architectures as reference, namely the UNet, UNETR, and SWinUNETR. We train them with the same losses and augmentations as the PUNet. In addition, we consider the established nnUNet framework \citep{isensee2021nnu}, which provides a robust set of extensive augmentations, a well-tested architecture and its own combination of Dice and focal losses. We calculate values for phase 1 (P1) and phase 2 (P2) where possible. For this experiment, P1 indicates a sole segmentation loss and P2 represents a self-supervised pre-training during P1 followed by full fine-tuning (non-frozen) in P2. Furthermore, we include the performance for 10\% of the original annotated training data. Results for all cases on the TCIA/BTCV and CT-ORG datasets are depicted in Table \ref{table_baselines}. The reported DSC values show that all variants of the PUNet, the prompted binary and multi class variants, as well as a non-prompted variant with a fixed linear output layer, achieve a comparable performance to the related architectures. The PUNet provides a similar performance as the SwinUNETR with a comparable parameter count. Interestingly, inserting prompts into the prompt-able blocks and relying on the aggregation scheme does not lead to a performance decrease, compared to the fixed architecture variant. We also see, that the nnUNet is superior in P1. This gap is less prevalent if its performance is compared to DSC values of P2 with e.g. a difference of $0.94$ percentage points (pp) between the nnUNet and the PUNet (binary). As expected, the benefit of the self-supervision scheme is more prominent when using only 10\% of the available training data, especially for TCIA/BTCV where an increase in DSC is seen regardless of the used architecture. This effect is less pronounced for the relatively larger CT-ORG dataset.

\begin{table*}[h!]
\centering
\footnotesize
\begin{tabular}{ lrrcrccr }
 \hline
 \multirow{2}{*}{Method} & \multirow{2}{*}{Parameters} & \multirow{2}{*}{Trainable} & \multicolumn{2}{c}{TCIA/BTCV} & & \multicolumn{2}{c}{CT-ORG} \\
 \cline{4-5} \cline{7-8}
 & & & DSC & p-value & & DSC & p-value \\
 \hline
 Fixed& $130$ & $0.00 \%$ & $24.52 \pm \phantom{0}5.07 \:\: (23.04)$ & $< 0.05$ & & $32.38 \pm \phantom{0}4.58 \:\: (32.98)$ & $< 0.05$\\
 Bias& $15\mathrm{k}$ & $0.23\%$ & $69.00 \pm 11.13 \:\: (70.69)$ & $< 0.05$ & & $78.87 \pm \phantom{0}7.03 \:\: (80.25)$ & $< 0.05$\\
 Prompting - w/o prompt bias scores & $44\mathrm{k}$ & $0.64 \%$ & $77.73 \pm \phantom{0}4.14 \:\: (77.50)$ & $< 0.05$ & & $85.30 \pm \phantom{0}6.87 \:\: (87.74)$ & $< 0.05$\\
 Prompting & $57\mathrm{k}$ & $0.85 \%$ & $79.62 \pm \phantom{0}3.81 \:\: (79.55)$ & $1.00$ & & $87.23 \pm \phantom{0}6.87 \:\: (88.42($ & $1.00$\\
 Bias + prompting& $73\mathrm{k}$ & $1.08\%$ & $79.94 \pm \phantom{0}4.10 \:\: (80.15)$ & $0.23$ & & $88.14 \pm \phantom{0}6.31 \:\: (90.01)$ & $< 0.05$\\
 Adapter& $325\mathrm{k}$ & $4.81\%$ & $81.07 \pm \phantom{0}4.69 \:\: (82.13)$ & $< 0.05$ & & $88.92 \pm \phantom{0}6.33 \:\: (90.52)$ & $< 0.05$\\
 Decoder& $2948\mathrm{k}$ & $43.63\%$ & $\mathbf{82.49 \pm \phantom{0}4.50 \:\: (83.20)}$ & $< 0.05$ & & $\mathbf{90.51 \pm \phantom{0}6.05 \:\: (92.25)}$ & $< 0.05$\\
 \hline
 Fine-tuning (non-frozen)& $6816\mathrm{k}$ & $100.85 \%$ & $83.13 \pm \phantom{0}3.86 \:\: (83.96)$ & $< 0.05$ & & $\mathbf{87.18 \pm \phantom{0}6.53 \:\: (89.91)}$ & $< 0.05$\\
 Fine-tuning (non-frozen) - fixed& $6758\mathrm{k}$ & $100.00 \%$ & $\mathbf{83.53 \pm \phantom{0}3.86 \:\: (83.99)}$ & $< 0.05$ & & $84.00 \pm \phantom{0}7.20 \:\: (85.32)$ & $< 0.05$\\
 \hline
\end{tabular}
\caption{Mean $\pm$ standard deviation (median) of the downstream segmentation performance of different adaptation schemes depicted by the Dice Similarity Coefficient (DSC, in \%) on the TCIA/BTCV and CT-ORG datasets. Statistical differences (p-values) are calculated between alternatives and our prompting variant. All shown approaches relied on the same jointly pre-trained backbone architecture, which is considered frozen during the downstream adaptation. The number of additional parameters is indicated in absolute values alongside the percentage with respect to the backbone architecture.}
\label{table_ablation_adaptations}
\end{table*}

\begin{figure*}[h]
    \centering
    \includegraphics[width=\textwidth]{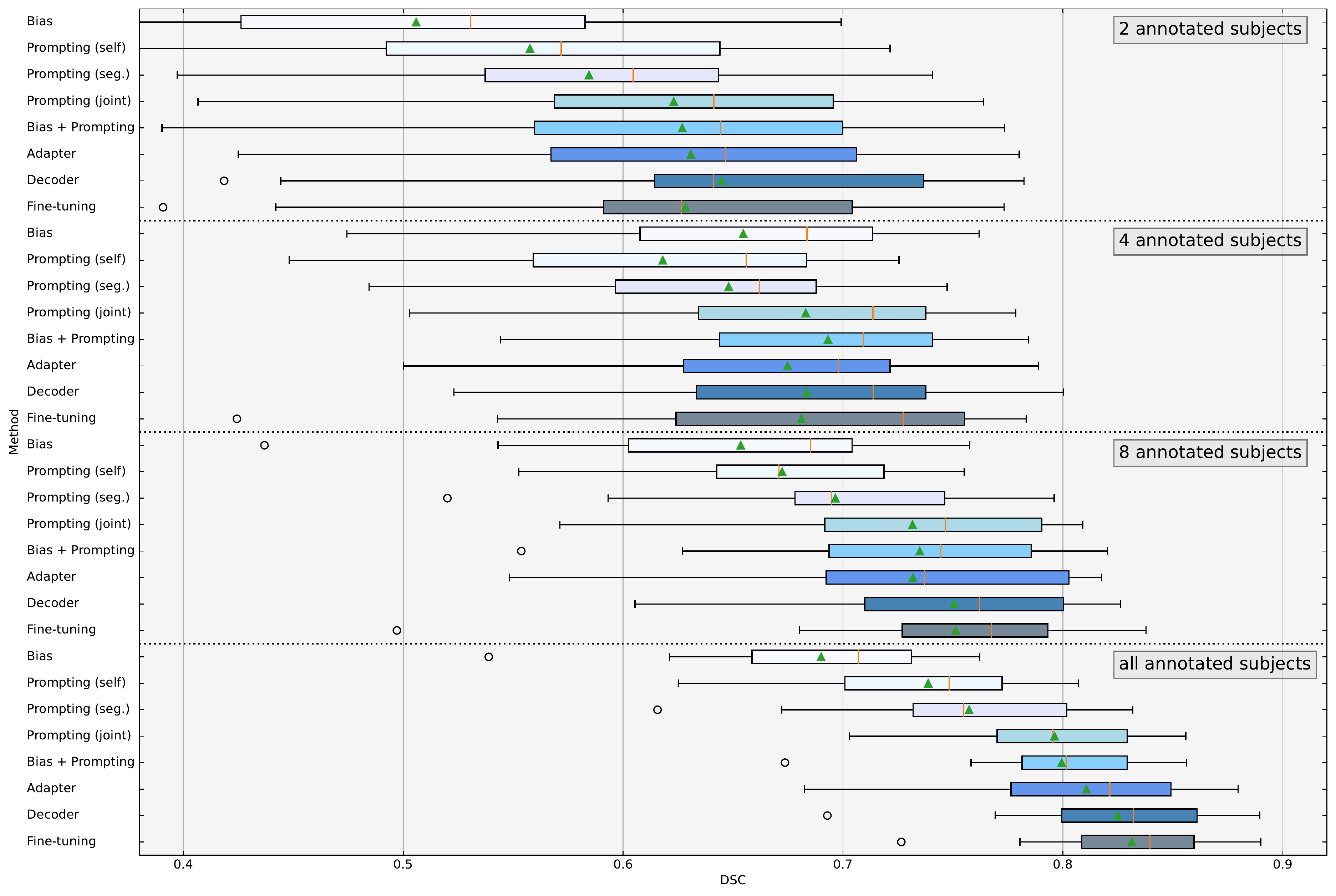}
    \caption{Boxplots depicting the downstream segmentation performance of adaptation variations by the Dice Similarity Coefficient (DSC, in \%) on the TCIA/BTCV dataset for varying amounts of available annotated subjects during the adaptation phase. The amount of annotated subjects (whole 3D volumes) includes 2, 4, 8 as well as the whole training set (60). Besides the differently pre-trained prompting variants (seg., self, joint), the alternatives make use of the jointly pre-trained network. Fine-tuning refers to the fully prompted and non-frozen variant. The boxplots include a mean (green diamond) and a median (orange line).} 
    \label{fig_scarcity}
\end{figure*}

In the second experiment, we investigate variations of the training schemes in P1 introduced in Section \ref{sec_training}. We evaluate the impact on metric values in P2. Note, that in this case all architectures in P2 can be either non-frozen for full fine-tuning or frozen for selective training. We consider four different training schemes for P1. As a baseline we introduce a random weight initialization with no actual pre-training performed. The other three variants are the proposed self-supervision scheme (\textit{self}) where no prompts are inserted into the prompt blocks in P1, pre-training by a sub-group of available annotations under use of class-conditional prompts in P1 with downstream adaptation on unseen classes in P2 (\textit{seg.}), and a combination of the class-conditional segmentation and self-supervision losses (\textit{joint}). For the latter case, pre-training and the respective adaptation is performed on the disjoint groups respectively. Mean DSC values for all cases are considered in Table \ref{table_training_schemes}. Naturally, the non-frozen models which allow for full parameter adaptation provide an upper baseline for the frozen counterparts. In the non-frozen case, there is little difference between the \textit{self}, \textit{seg.}, and \textit{joint} schemes. There is a performance drop-off of $1.24$ pp for TCIA/BTCV and $4.88$ pp for CT-ORG when comparing the random baseline to the best variants. For a frozen model however, a good initialization becomes mandatory, with the self-supervision and segmentation pre-training both providing substantial increases in DSC values above the random initialization. The combination of both pre-training schemes is even more beneficial and is $3.83$ pp and $2.67$ pp below the non-frozen models. The ASSD values, indicating deviations from the predicted segmentation mask boundary to the ground truth, shows a similar behavior to the DSC values, with lowest values achieved for the \textit{joint} scheme in the frozen case. Yet, they differ for the CT-ORG data. This is only the case due to a skewed distribution, since the median value $2.68$ is still well below the value of the self variant of $3.76$. In all cases, we see, that even the random initialization is adaptable to new classes by the inserted prompt tokens, albeit with a larger performance decrease.

As seen in the previous experiments, we establish the combination of pre-training and prompting as a valid scheme for a downstream adaptation to unseen classes. Next, we want to identify how effective it is in operating on a frozen model compared to established approaches. As such, we consider a series of common adaptation approaches and report results alongside the amount of trainable parameters in Table \ref{table_ablation_adaptations}. All ablations use the same PUNet backbone architecture, with the self-supervision and segmentation pre-training applied in P1. We consider the following ablations: a trainable linear output layer (fixed), trainable bias and normalization parameters (bias) and the combination of bias and prompting (bias + prompting), prompting without additional attention bias scores for prompt tokens, an additional linear layer (including normalization and activation) at the end of each prompt-able block (adapter), a non-frozen (fully trainable) decoder (decoder), as well as full fine-tuning for the prompted and non-prompted architecture. Results are reported with respect to DSC values and a p-value indicating statistical differences between the respective variant and our proposed prompting scheme. A fixed layer incurs the least parameter adjustments (130), followed by bias adjustments (15k), prompting (57k) and the adapter variant (325k). All require less than 5\% of parameters to be adjusted compared to a full fine-tuning. The decoder already requires a large amount of parameter adjustments of 43.63\%. For the performance, we see that full fine-tuning as well as the decoder variant are superior to parameter efficient schemes. With decreasing amounts of parameters the performance gradually continues to worsens with substantial drop-offs for the bias and fixed variants. This trend is present for TCIA/BTCV and CT-ORG alike. The single fixed layer, is not sufficient to provide adequate adaptation. Yet, our prompting scheme is only $2.87$ pp and $3.28$ pp in DSC below the decoder adaptation while requiring only $1.93\%$ of its amount of parameter. It is also close in performance to the adapter variant with a difference of $1.45$ and $1.69$ pp in DSC with $17.5\%$ of the amount of its parameters. All results are statistically significant, except for the prompting and bias + prompting variant on TCIA/BTCV, where the differences are minor. 

To further deepen our understanding of the efficacy of the adaptation schemes, we consider the case of limited annotated training data during the downstream task in P2. This does not affect P1, where the whole training data and respective masks (of seen classes) are still available for the self- and segmentation supervision. However, in P2 only few annotated subjects for the new unseen classes are made available. We consider a subset of the variants introduced in the previous experiment. In addition, we include a prompting variant with only self-supervision and a variant with only segmentation supervision. We vary the number of available annotations from 2 subjects to 4, 8, and the whole TCIA/BTCV dataset. To relate the amounts to the first experiment, 2 subjects correspond to $3.3\%$, 4 subjects to $6.7\%$ and 8 subjects to $13.3\%$ of the whole training dataset. The boxplots are depicted in Figure \ref{fig_scarcity}. The overall order of the performance of different schemes seen in Table \ref{table_ablation_adaptations} remains mostly intact for lower amounts of annotated data. The combination of self supervision and segmentation pre-training is superior to each of the losses alone as well as the bias adaptation for all amounts. This variant is also slightly better than the adapter approach for 4 and 8 subjects. As before, there is little benefit in applying bias adaptations in combination with the prompt tuning. Full fine-tuning is superior in most cases, however, for 2 and 4 subjects, decoder only training lead to the best adaptation. To a lesser degree, this aspect is also present for the comparison of our joint scheme and the full fine-tuning, with both performing similarly for 2 and 4 subjects.

For architectural ablations, we considered variants of the final similarity aggregation scheme. We differentiate between the weighted aggregation, as proposed in Section \ref{sec_aggregation}, a top-$k$ selection with $k=3$, indicating that the three most similar prompt token vectors are considered eligible, and a simple mean aggregation, where the similarity of all token vectors of a class is simply averaged to identify its overall similarity score. In addition, we investigated the impact of amounts of tokens used to represent a class. Hereby, we vary the amount from a single token to 32 tokens. Note, that the amount of tokens also influences the amount of parameters, that can be adjusted.  Investigations are performed on the TCIA/BTCV dataset and depicted in Table \ref{table_ablation_aggregation}. The proposed scheme is slightly ahead, followed by top-$k$ and the mean aggregation. However, the differences are statistically insignificant as shown by a high p-value ($> 0.05$). With each token the performance drastically increased with respect to DSC as well as ASSD. This trend diminishes for higher amounts of tokens. For 16 and 32 tokens, the difference became insignificant and the DSC and ASSD values stagneted.

Further, we can vary the selection of PSWin blocks at which prompts are inserted into the network. It is possible, to use deep prompting throughout the architecture (ours, full), to only rely on the final similarity aggregation (sim. agg.), which is similar to earlier prototype networks, to consider only the first prompt block (and in this work the final aggregation), to make it more akin to large language model prompting (start), or to rely only on the final aggregation in combination with prompt-able blocks adjusted solely in the encoder or decoder. The impact of the placement is reported by DSC and p-values in Table \ref{table_ablation_prompting}. As expected, deep prompting performs best, with the decoder only variant following with a decrease of $2.8$ pp in DSC. This is more important, than early prompting in the architecture. The ablation also shows, that merely adding prompt tokens to the first prompt-able block in the encoder and the final aggregation is not sufficient to achieve adaptation to an unseen class.

\begin{table}[t!]
\centering
\footnotesize
\begin{tabular}{ lccr }
 \hline
 Method & DSC & ASSD & p-value \\
 \hline
 Weighted agg. (std.) & $\mathbf{79.62 \pm \phantom{0}3.81}$ & $\mathbf{2.30 \pm 1.14}$ & $1.00$\\
 top-$k$ agg. & $79.34 \pm \phantom{0}3.72$ & $2.31 \pm 1.10$ & $ 0.11$\\
 Mean agg. & $79.23 \pm \phantom{0}4.62$ & $2.38 \pm 1.57$ & $ 0.34$\\
 \hline
 $T=1$ & $52.19 \pm \phantom{0}5.11$ & $7.47 \pm 3.14$ & $< 0.05$\\
 $T=2$ & $70.22 \pm \phantom{0}4.05$ & $3.85 \pm 1.76$ & $< 0.05$\\
 $T=4$ & $74.82 \pm \phantom{0}4.15$ & $3.01 \pm 1.66$ & $< 0.05$\\
 $T=8$ & $76.85 \pm \phantom{0}4.29$ & $2.71 \pm 1.83$ & $< 0.05$\\
 $T=16$ (std.) & $\mathbf{79.62 \pm \phantom{0}3.81}$ & $\mathbf{2.30 \pm 1.14}$ & $1.00$\\
 $T=32$ & $79.51 \pm \phantom{0}4.16$ & $2.47 \pm 1.72$ & $0.71$\\
 \hline
\end{tabular}
\caption{Mean $\pm$ standard deviation (median) of the Dice Similarity Coefficient (DSC, in \%) and Average Symmetric Surface Distance (ASSD, in mm) and p-values with respect to the standard (std.) variant for different ablations during the downstream adaptation on the TCIA/BTCV dataset. We compared the proposed weighted similarity aggregation (weighted agg.), a top-$k$ selection among available class tokens with $k=3$ (top-$k$), and a mean aggregation (mean agg.) where similarities are averaged across all tokens of a class. In addition, the amount of tokens $T$ per class within every prompt-able block and within the final similarity aggregation is varied between 1 and 32.}
\label{table_ablation_aggregation}
\end{table}

\begin{table}[t!]
\centering
\footnotesize
\begin{tabular}{ lrcc }
 \hline
 Method & Parameters & DSC & ASSD \\
 \hline
 Full & $57\mathrm{k}$ & $\mathbf{79.62 \pm \phantom{0}3.81}$ & $\mathbf{\phantom{0}2.30 \pm \phantom{0}1.14}$\\
 Sim. agg. & $675$ & $\phantom{0}1.58 \pm \phantom{0}0.85$ & $53.08 \pm 26.36$ \\
 Start (+ sim. agg.). & $2\mathrm{k}$ & $\phantom{0}5.58 \pm \phantom{0}1.11$ & $20.39 \pm \phantom{0}9.17$ \\
 Encoder (+ sim. agg.). & $37\mathrm{k}$ & $70.37 \pm \phantom{0}5.23$ & $\phantom{0}3.34 \pm \phantom{0}1.35$ \\
 Decoder (+ sim. agg.). & $21\mathrm{k}$ & $76.82 \pm \phantom{0}4.59$ & $\phantom{0}2.85 \pm \phantom{0}1.86$ \\
 \hline
\end{tabular}
\caption{Mean $\pm$ standard deviation (median) of the downstream segmentation performance of different prompt insertion variants depicted by the Dice Similarity Coefficient (DSC, in \%) and the Average Symmetric Surface Distance (ASSD, in mm) on the dataset alongside the amounts of prompt token (and attention bias score) parameter required for the adaptation.}
\label{table_ablation_prompting}
\end{table}

\section{Discussion}
The proposed architecture enables the insertion of prompt tokens throughout the whole segmentation network. This is not possible with classical networks in the medical field which relied predominantly on convolutions, which are inherently limited in their ability to process heterogeneous content in an efficient manner. Despite the comparatively small number of available model and even fewer prompt parameters, our contribution shows great efficacy in the adaptation to unseen classes on two CT datasets with varying organs and structures. The underlying architecture delivers comparable performances to the recent SwinUNETR. This is the case for binary, multi-class and fixed predictions alike, despite the PUNet operating on half the resolution and thus relying on a simple final upsampling layer. No architectural alterations have to be performed when switching the prompts. Thereby, several segmentation targets can be trained on concurrently in a single batch (on frozen and non-frozen models). We show that prompting is effective at adapting a frozen pre-trained model in a downstream fine-tuning adaptation, significantly closing the performance gap between full fine-tuning and prompt tuning. The heterogeneous bias score further aids in adjusting the network to unseen classes. We find that a certain number of prompt tokens for each class suffice with further tokens leading to diminishing returns. We present a token dependent aggregation scheme that can replace a fixed output layer.

The self-supervision pre-training scheme with its online generated prototypes leads to a robust embedding, where anatomical regions can be distinguished from each other. This is the case in spite of heavy masking or the presence of a concurrent prompting objective. As can be seen, in Figure \ref{fig_data} and \ref{fig_dimensions} the pre-training scheme not only allows for enhanced performance in the downstream task, but would also be usable for landmark localization and could be further strengthened for this task by adjustment of the FWHM parameter in the online clustering. It also shows a beneficial increase in DSC values across different architectures. The benefit of including segmentation masks directly in the pre-training is shown by significantly increased DSC and reduced ASSD values. For TCIA/BTCV the segmentation pre-training is even more valuable than the self-supervision. With a segmentation pre-training, the network is able to learn to incorporate prompts in a more efficient way and is also acquainted with delineating borders for sharp mask predictions as can be seen by lower ASSD values. The introduction of our self-supervision as well as the segmentation supervision is most beneficial in the joint application. The pre-training strategies are as such complementary. The effect of applying a pre-training scheme becomes necessary, when working with frozen models in P2. As such, the gap between full fine-tuning and frozen model adaptation is significantly lessened.

The adaptation ablation experiments comparing prompting to common adaptation schemes establish a trend where more trainable parameters correlate with a higher DSC value. This trend however, achieves diminishing returns for increasing amounts of parameters. Under this aspect, the prompting scheme proofs especially flexible, since the prompts are not part of the backbone network architecture. The ablation of the prompt insertion indicates, that adjustments throughout the decoder are most important to incur meaningful changes in the output embedding, so that the aggregation can identify pixels of a certain class properly. This is in line with the recent Segmenter architecture \citep{strudel2021segmenter}, which relied on a joint processing of class tokens and encoded image content in its decoder.

The scope of this work remains limited. We focus on the impact of the introduced training schemes with respect to their efficacy for different downstream adaptation strategies. We do not evaluate the pre-training strategies against state-of-the-art approaches of each respective sub-field and do not claim superiority. We do not compare our method against the vast number of alternatives schemes in the label scarce case known from the semi-supervised literature \cite{yang2021survey}. In addition, the label scarcity could be extended to multiple seeds for better statistical power. The proposed approach is a mere first step with a multitude of potential extensions. The architecture could operate on full resolution inputs at the cost of higher memory consumption and training time. As shown for language models by \cite{brown2020language}, we expect that greatly increasing the neural network model parameters can have strong beneficial effects on the label scarce performance and may also positively influence the requirements on the amount of prompt token parameters and additive bias scores. In addition, neighborhood attention \citep{hassani2022neighborhood, hassani2022dilated} has been proposed recently to avoid the use of potentially restrictive shifted windows entirely. More sophisticated hierarchical clustering could be employed for the online prototype generation, e.g. on multiple resolution levels. Likewise, a more sophisticated prompt token generation could be introduced, to further reduce the amount of stored parameters, as done by \cite{he2022hyperprompt}. We note that when relying on the binary case, an implicit extension to incremental class learning can be achieved circumventing catastrophic forgetting entirely. Further investigations are to be performed for the combination of prompts from different known classes or the introduction of a new class in the multi-class case. Hereby, the model needs to be able to adjust prompt tokens to the new circumstances while not forgetting to predict known classes.

\section{Conclusion}
In this work, we propose advancements in the efficacy and applicability of using pre-trained frozen models for downstream segmentation tasks based on a prompt tuning scheme. We establish a scheme and architecture which provides natural insertion points for task-dependent tuning, while leaving the original pre-trained model intact. Introducing several pre-requisites, we make a step towards closing the performance between non-frozen and frozen models, allowing for readily re-usable models in the field. In light of the results, further investigations, e.g. in the composability of prompt tokens of different classes seem promising and interesting avenues for future endeavours.

\bibliographystyle{elsarticle-harv.bst}\biboptions{authoryear}
\bibliography{references}

\end{document}